%% file: neurips_2021.tex
\documentclass{article}

\PassOptionsToPackage{numbers, compress}{natbib}

\usepackage[preprint]{neurips_2021}

\usepackage{graphicx}
\usepackage{caption}
\usepackage[utf8]{inputenc} 
\usepackage[T1]{fontenc}    
\usepackage[pdfencoding=auto]{hyperref}     
\usepackage{url}            
\usepackage{booktabs}       
\usepackage{amsmath}
\usepackage{multirow}
\usepackage{multicol}
\usepackage{amsfonts}       
\usepackage{nicefrac}       
\usepackage{microtype}      
\usepackage{algorithm}
\usepackage{algorithmic}
\usepackage{xcolor}         
\usepackage{caption} 
\usepackage{subcaption}
\usepackage{comment}
\usepackage{listings}
\usepackage{hyperref}
\hypersetup{colorlinks,linkcolor={blue},citecolor={green},urlcolor={blue}}  

\DeclareMathOperator{\Tr}{Tr}
\newtheorem{theorem}{Theorem}
\newcommand\numberthis{\addtocounter{equation}{1}\tag{\theequation}}

\definecolor{codegreen}{rgb}{0,0.6,0}
\definecolor{codegray}{rgb}{0.5,0.5,0.5}
\definecolor{codepurple}{rgb}{0.58,0,0.82}
\definecolor{backcolour}{rgb}{0.95,0.95,0.92}
\lstdefinestyle{mystyle}{
    inputencoding=utf8,
    extendedchars=true,
    commentstyle=\color{codegreen},
    keywordstyle=\color{magenta},
    numberstyle=\tiny\color{codegray},
    stringstyle=\color{codepurple},
    basicstyle=\ttfamily\small,
    breakatwhitespace=false,         
    breaklines=true,                 
    captionpos=b,                    
    keepspaces=true,                 
    numbers=left,                    
    numbersep=5pt,                  
    showspaces=false,                
    showstringspaces=false,
    showtabs=false,                  
    tabsize=4,
    literate={λ}{{$\lambda$}}1
}
\title{\texttt{SpreadGNN}: Serverless Multi-task Federated Learning for Graph Neural Networks}

\author{%
  Chaoyang He \thanks{Equal contribution (alphabetical order). Email: \texttt{\{chaoyang.he,ceyani,keshavba\}@usc.edu}.} , Emir Ceyani \footnotemark[1] , Keshav Balasubramanian \footnotemark[1] \And Murali Annavaram \& Salman Avestimehr \\
  Viterbi School of Engineering\\
  university of Southern California
}

\begin{document}

\maketitle

\begin{abstract}
Graph Neural Networks (GNNs) are the first choice methods for graph machine learning problems thanks to their ability to learn state-of-the-art level representations from graph-structured data. However, centralizing a massive amount of real-world graph data for GNN training is prohibitive due to user-side privacy concerns, regulation restrictions, and commercial competition. Federated Learning is the de-facto standard for collaborative training of machine learning models over many distributed edge devices without the need for centralization. Nevertheless, training graph neural networks in a federated setting is vaguely defined and brings statistical and systems challenges. This work proposes \texttt{SpreadGNN}, a novel multi-task federated training framework capable of operating in the presence of partial labels and absence of a central server for the first time in the literature. \texttt{SpreadGNN} extends federated multi-task learning to realistic serverless settings for GNNs, and utilizes a novel optimization algorithm with a convergence guarantee, \textit{Decentralized Periodic Averaging SGD (DPA-SGD)}, to solve decentralized multi-task learning problems. We empirically demonstrate the efficacy of our framework on a variety of  non-I.I.D. distributed graph-level molecular property prediction datasets with partial labels. Our results show that  \texttt{SpreadGNN} outperforms GNN models trained over a central server-dependent federated learning system, even in constrained topologies. The source code is publicly available at \url{https://github.com/FedML-AI/SpreadGNN}.
\end{abstract}

\input{sections/intro.tex}

\input{sections/problem.tex}
\input{sections/exps.tex}
\input{sections/related.tex}

\input{sections/conclusion.tex}

\bibliographystyle{plainnat} 
\bibliography{FMTDL}

\clearpage
\section*{Appendix}

This section includes additional information that a reader might find useful. Apart from the proof of Theorem \ref{DPA-SGD-theorem}, we include the algorithm sketch for SpreadGNN using DPA-SGD, a more detailed description of the datasets we used, hyperparameter configurations used in our experiments and additional ablation studies on communication period and network topology.  

\input{sections/appendix-dataset_details.tex}
\input{sections/appendix-hparams.tex}

\input{sections/appendix-ablation.tex}

\input{sections/appendix-proof.tex}

\end{document}

%% file: sections/intro.tex
\section{Introduction}
\label{sec:intro}

Graph Neural Networks (GNNs) \cite{ graphsage} are expressive models that can distill structural knowledge into highly representative embeddings. While graphs are the representation of choice in domains such as social networks[3,31], knowledge graphs for recommendation systems \citep{chen2020survey}, 
in this work we focus on molecular graphs that are the core of drug discovery, molecular property prediction [13, 23] and virtual screening [56, 64]. Molecular graphs differ from their more well known counterparts such as social network graphs. First, each molecule is a graph representation of the basic atoms and bonds that constitute the molecule and hence the size of the graph is small. Second, even though each graph may be small there are numerous molecules that are being developed continuously for varied use cases. Hence, what they lack in size they make up for it in structural heterogeneity. Third, molecules can be labeled along multiple orthogonal dimensions.  Since each graph has multiple labels the learning itself can be characterized as multi-task learning. For instance, whether a molecule has potentially harmful interaction with a diabetics drug, or whether that molecule can turn toxic under certain conditions are distinct labels. Molecular property analysis and labeling requires wet-lab experiments, which is time-consuming and resource-costly. As a consequence, many entities may only have partially labeled molecules even if they know the graph structure. Finally, molecules are coveted inventions and hence entities often possess proprietary graph representation that cannot be shared with other institutions for competitive and regulatory reasons. But training collectively over a private set of molecular graphs can have immense societal benefits such as accelerated drug discovery.   

\begin{figure*}[t]
  \centering
  \includegraphics[width=.7\linewidth]{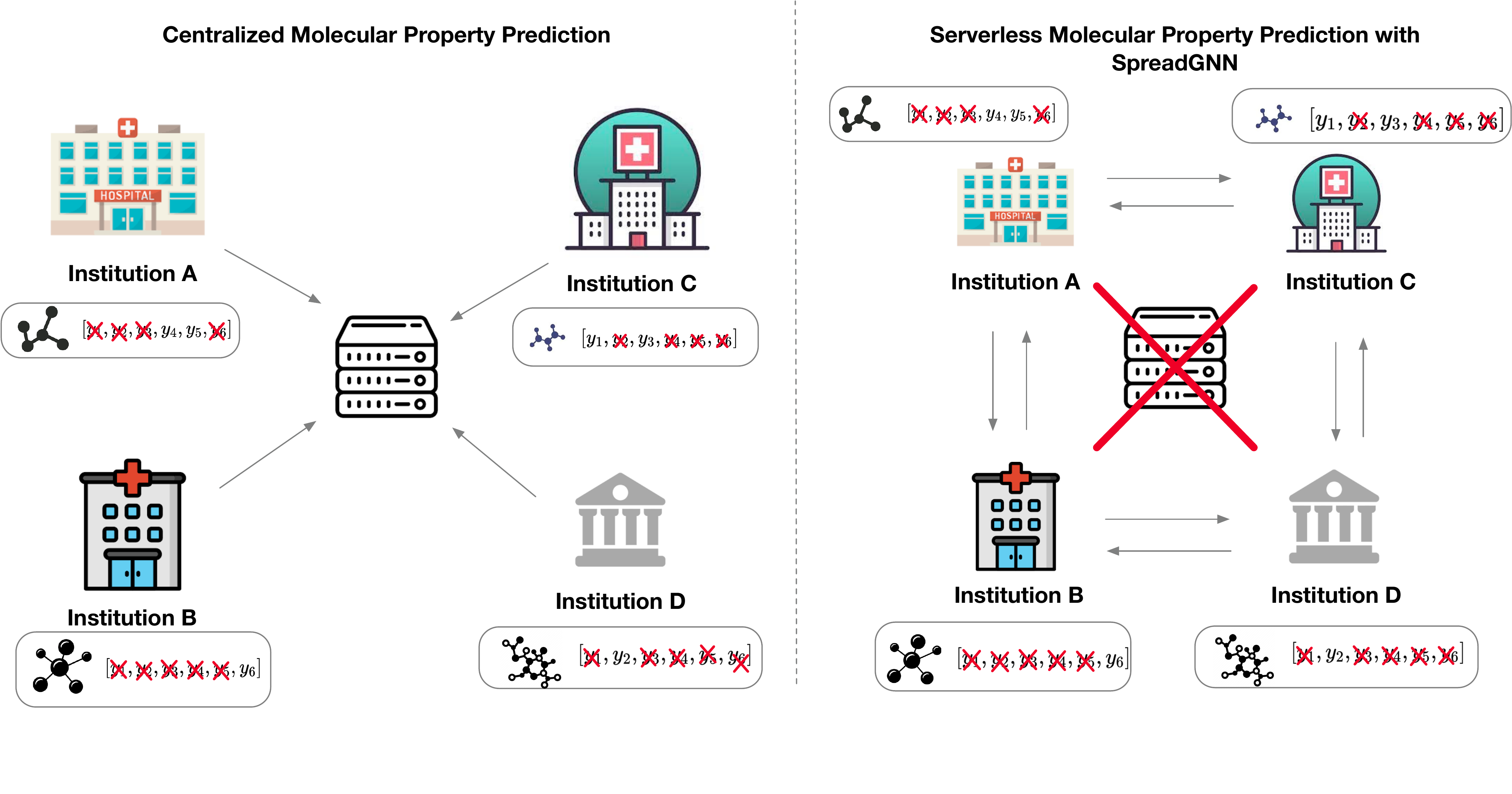}
  \caption{\textit{Serverless Multi-task Federated Learning for Graph Neural Networks}.}
  \label{fig:framework}
\vspace{-0.1in}
\end{figure*}

Federated Learning (FL) is a distributed learning paradigm that addresses this data isolation problem via collaborative training. In this paradigm, training is an act of collaboration between multiple clients (such as research institutions) without requiring centralized local data while providing a certain degree of user-level privacy \citep{mcmahan2017communication, kairouz2019advances}. However there are still challenges and shortcomings to training GNNs in a federated setting. As shown in \citep{fedgraphnn}, federated GNNs perform poorly in a non-iid setting. This setting (Figure \ref{fig:framework}) is the typical case in molecular graphs since each owner may have different molecules and even when they have the same molecular graph each owner may have an incomplete set of labels for each molecule. The left half of Figure \ref{fig:framework} shows a simpler case where all the clients can communicate through a central server. But in practice the presence of a central server is not feasible when multiple competing entities may want to collaboratively learn. The challenges are further compounded by the lack of a central server as shown in the right half of the Figure \ref{fig:framework}. Thus, it remains an open problem to design a federated learning framework for molecular GNNs, for a realistic setting, in which clients only have partial labels and one in which there is no reliance on a central server. This is the problem we seek to address in this work. 

We propose a multitask federated learning framework called  \texttt{SpreadGNN} that operates in the presence of partial labels and absence of a central server as shown in Figure \ref{fig:framework}. We use the word task and class interchangeably, with each label being composed of multiple tasks. First, we present a multitask learning (MTL) formulation to learn from partial labels. Second, in our MTL formulation, we utilize decentralized periodic averaging stochastic gradient descent (DPA-SGD) to solve the serverless MTL optimization problem, and also provide a theoretical guarantee on the convergence properties for DPA-SGD, which further verifies the rationality of our design.

We evaluate \texttt{SpreadGNN} on graph level molecular property prediction and regression tasks. We synthesize non-I.I.D. and partially labeled datasets by using curated data from the MoleculeNet \citep{wu2018moleculenet} machine learning benchmark. With extensive experiments and analysis, we find that \texttt{SpreadGNN} can achieve even better performance than \texttt{FedAvg} \citep{fedavg}, not only when all clients can communicate with each other, but also when clients are constrained to communicate with a subset of other clients. We plan on publishing the source code of \texttt{SpreadGNN} as well as related datasets for future exploration.

%% file: sections/problem.tex
\section{\texttt{SpreadGNN} Framework}
\subsection{Preliminary: Federated Graph Neural Networks for Graph Level Learning}

\label{sec:formulation}

We seek to learn \textit{graph level} representations in a federated learning setting  over decentralized graph datasets located in edge servers which cannot be centralized for training due to \textbf{privacy and regulation restrictions} \cite{fedgraphnn}. For instance, compounds in molecular trials \citep{rong2020self} may not be shared across entities because of intellectual property or regulatory concerns. 
Under this setting, we assume that there are $K$ clients in the FL network, and the $k^{th}$ client has its own dataset $\mathcal{D}^{(k)}:=\left\{\left(G_{i}^{(k)}, y_{i}^{(k)}\right)\right\}_{i=1}^{N^{(k)}}$, where $G_{i}^{(k)}=(\mathcal{V}_{i}^{(k)}, \mathcal{E}_{i}^{(k)})$ is the $i^{th}$ graph sample in $\mathcal{D}^{(k)}$ with node \& edge feature sets $\boldsymbol{X}^{(k)} = \left\{ \boldsymbol{x}_{m}^{(k)}\right\}_{m \in \mathcal{V}_{i}^{(k)}}$ and $\boldsymbol{Z}^{(k)} = \left\{\boldsymbol{e}_{m, n}^{(k)}\right\}_{m, n \in \mathcal{V}_{i}^{(k)}}$, $\mathbf{y_{i}}^{(k)}$ is the corresponding multi-class label of $G_{i}^{(k)}$,  $N^{(k)}$ is the sample number in dataset $\mathcal{D}^{(k)}$, and $N = \sum_{k=1}^{K} N^{(k)}$. 

Each client owns a GNN with a readout, to learn graph-level representations. We call this model of a GNN followed by a readout function, a \textit{graph classifier}. 
Multiple clients are interested in collaborating  to improve their GNN models without necessarily revealing their graph datasets. In this work, we build our theory upon the Message Passing Neural Network (MPNN) framework \citep{gilmer2017neural,10.1145/3394486.3406474} as most spatial GNN models \citep{kipfgcn,gat,graphsage} can be unified into this framework. The forward pass of an MPNN  has two phases: a message-passing phase (Eq \eqref{message}) and an update phase (Eq \eqref{passing}). For each client, we define the graph classifier with an L-layer GNN followed by a readout function as follows:
\begin{gather}
\boldsymbol{m}_{i}^{(k, \ell +1)} =\operatorname{\text{AGG}}\left(\left\{\boldsymbol{M}_{\mathbf{\theta}}^{(k, \ell +1)}\left(\boldsymbol{h}_{i}^{(k, \ell)}, \boldsymbol{h}_{j}^{(k, \ell)}, \boldsymbol{e}_{i, j}\right) \mid j \in \mathcal{N}_{i}\right\}\right), \ell = 0,\dots , L-1 \label{message}\\
\boldsymbol{h}_{i}^{(k, \ell +1)} =\boldsymbol{U}_{\mathbf{\Psi}}^{(k, \ell +1)}\left(\boldsymbol{h}_{i}^{(k, \ell)}, \boldsymbol{m}_{i}^{(k, \ell +1)}\right) , \ell = 0,\dots , L-1 \label{passing} \\
\hat{\mathbf{y}}_{i}^{(k)}=\boldsymbol{R}_{\mathbf{\Phi_{\text{pool}}}, \mathbf{\Phi_{\text{task}}}}\left(\left\{h_{j}^{(k, L)} \mid j \in \mathcal{V}_{i}^{(k)}\right\}\right)  \label{readout}
\end{gather} where $\boldsymbol{h}^{(k, 0)}_i = \boldsymbol{x}_{i}^{(k)}$ is the $k^{th}$ client's node features, $\ell$ is the layer index, $\texttt{AGG}$ is the aggregation function (e.g., in the GCN model \citep{kipfgcn}, the aggregation function is a simple $\texttt{SUM}$ operation), and $\mathcal{N}_{i}$ is the neighborhood set of node $i$. In Eq. \eqref{message}, $\boldsymbol{M}_{{\theta}^{(k, \ell +1)}}\left( \cdot \right)$ is the message generation function which takes the hidden state of current node $\boldsymbol{h}_{i}$, the hidden state of the neighbor node $\boldsymbol{h}_{j}$ and the edge features $\boldsymbol{e}_{i, j}$ as inputs to gather and transform neighbors' messages. In other words, $\boldsymbol{M}_{\theta}$ combines a vertex's hidden state with the edge and vertex data from its neighbors to generate a new message.  $\boldsymbol{U}_{\Psi}^{(k, \ell+1)}\left( \cdot \right)$ is the state update function that updates the model using the aggregated feature $\boldsymbol{m}_{i} ^{(k, \ell + 1)}$ as in Eq. \eqref{passing}. After propagating through $L$ GNN layers, the final module of the graph classifier is a readout function $\boldsymbol{R}_{\mathbf{\Phi_{\text{pool}}}, \mathbf{\Phi_{\text{task}}}}\left( \cdot \right)$ which allows clients to predict a label for the graph, given node embeddings that are learned from Eq.\eqref{passing}. In general the readout is composed of two neural networks: the pooling function parameterized by $\mathbf{\Phi}_{\text{pool}}$; and a task classifier parameterized by $\mathbf{\Phi}_{\text{task}}$. The role of the pooling function is to learn a single graph embedding given node embedding from Eq \eqref{passing}. The task classifier then uses the graph level embedding to predict a label for the graph.

To formulate GNN-based FL, using the model definition above, we define $\boldsymbol{W} = \left\{\mathbf{\theta}, \mathbf{\Psi} , \mathbf{\Phi}_{\text{pool}} , \mathbf{\Phi}_{\text{task}} \right\}$
as the overall learnable weights. Note that $\boldsymbol{W}$ is independent of graph structure as both the GNN and Readout parameters make no assumptions about the input graph. Thus, one can learn $\boldsymbol{W}$ using a FL based approach. The the overall FL task can be formulated as a distributed optimization problem as:
\begin{equation}
\small{\min_{\boldsymbol{W}} F(\boldsymbol{W}) \stackrel{\text {def}}{=} \min_{\boldsymbol{W}} \sum_{k=1}^{K} \frac{N^{(k)}}{N} \cdot f^{(k)}(\boldsymbol{W}) \label{eq:FL}}
\end{equation}
where  $\space f^{(k)}(\boldsymbol{W}) = \frac{1}{N^{(k)}} \sum_{i=1}^{N^{(k)}} \mathcal{L}( \mathbf{\hat{y}}_{i}^{(k)}, \mathbf{y}_{i}^{(k)})$ is the $k^{th}$ client's local objective function that measures the local empirical risk over the heterogeneous graph dataset $\mathcal{D}^{(k)}$. $\mathcal{L}$ is the loss function of the global graph classifier. 

With such a formulation, it might seem like an optimization problem tailored for a \texttt{FedAvg} based optimizers \citep{fedavg, he2020group, reddi2020adaptive}. Unfortunately, in molecular graph settings this is not the case for the following reasons. (a) In our setting, clients belong to a decentralized, serverless topology. There is no central server that can average model parameters from the clients. (b) Clients in our setting possess incomplete labels, hence the dimensions of $\mathbf{\Phi_{task}}$ can be different on different clients. For instance, a client may only have a partial toxicity labels for a molecule, while another client may have a molecule's interaction property with another drug compound. Even with such incomplete information from each client, our learning task is interested in classifying each molecule across multiple label categories. Federated Multitask Learning (FMTL) \citep{smith_federated_2017, caldas_federated_2018} is a popular framework designed to deal with such issues. However, current approaches do not generalize to non-convex deep models such as graph neural networks. To combat these issues, we first introduce a centralized FMTL framework for graph neural networks in Section \ref{sec:fedgmtl}. But reliance on a central server may not be feasible in molecular graphs owned by multiple competing entities. Hence, we enhance this centralized FMTL to a serverless scenario in section \ref{sec:de-gmtl},  which we call  \texttt{SpreadGNN}.

\subsection{Federated Multi-Task Learning with Graph Neural Networks} 
\label{sec:fedgmtl}

\begin{figure*}[h]
  \centering
  \includegraphics[width=\linewidth]{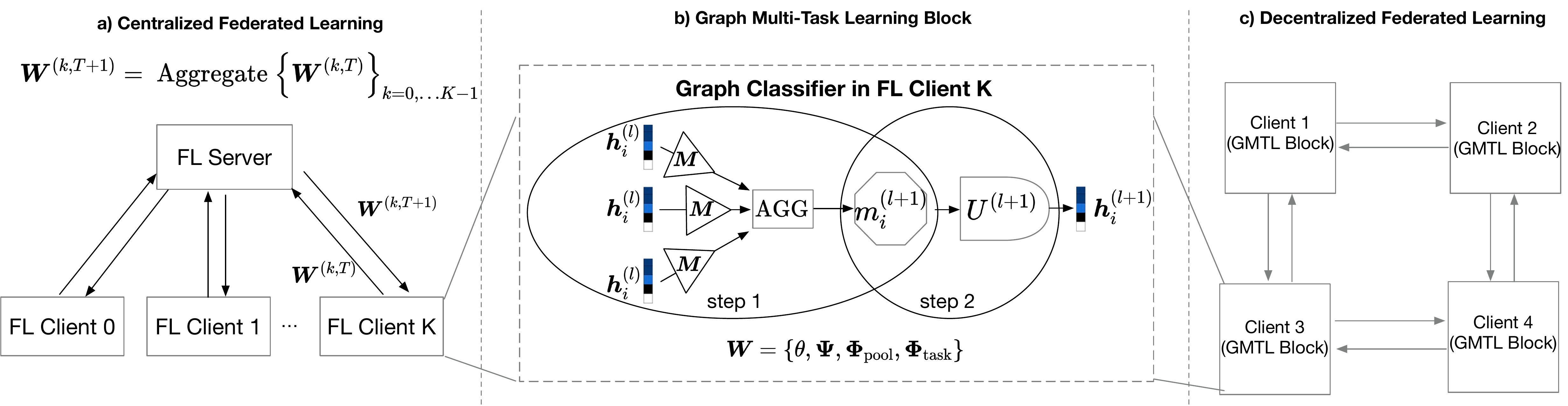}
  \caption{\textit{Federate Graph MultiTask Learning Framework (FedGMTL)}.}
  \label{fig:framework2}
\end{figure*}

Under the regularized MTL paradigm \cite{evgeniou2004regularized}, we define the centralized federated graph MTL problem (FedGMTL) as follows: 

\begin{equation}
\label{eq:fedgmtl}
\begin{aligned}
& \begin{aligned}[t] \min_{\mathbf{\theta}, \mathbf{\Psi} , \mathbf{\Phi}_{\text{pool}} , \mathbf{\Phi}_{\text{task}} }
&\sum_{k=1}^{K}\frac{1}{N_k} \sum_{i=1}^{N_k}\mathcal{L}( \mathbf{\hat{y}}_{i}^{(k)}, \mathbf{y}_{i}^{(k)}) ) + \mathcal{R}(\boldsymbol{W}^{(k)},\mathbf{\Omega}) , \end{aligned} \\
& \qquad \text{s.t.} \qquad  \mathbf{\Omega} \ge 0 \quad \text{and} \quad \Tr(\mathbf{\Omega}) = 1. 
\end{aligned}
\end{equation}
where
\begin{equation}
\label{eq:eq6}
    \mathcal{R}(\boldsymbol{W}^{(k)},\mathbf{\Omega}) = \frac{1}{2}\lambda_1 \Tr(\mathbf{\Phi}_{\text{task}} \mathbf{\Omega}^{-1} \mathbf{\Phi}^{T}_{\text{task}}) + \frac{1}{2} \sum_{ \boldsymbol{\chi} \in \left\lbrace   \mathbf{\theta}, \mathbf{\Psi} , \mathbf{\Phi}_{\text{pool}} , \mathbf{\Phi}_{\text{task}} \right\rbrace }\lambda_{\boldsymbol{\chi} } ||\boldsymbol{\chi}||_F^2
\end{equation} is the bi-convex regularizer introduced in \cite{zhang_convex_2012}. The first term of the Eq. \eqref{eq:fedgmtl}  models the summation of different empirical loss of each client, which is what Eq. \eqref{eq:FL} exactly tries to address. The second term serves as a task-relationship regularizer with $\mathbf{\Omega} \in \mathbb{R}^{ \mathcal{S} \times \mathcal{S} }$ being the covariance matrix for $\mathcal{S}$ different tasks constraining the task weights $\mathbf{\Phi}_{\text{task}} = [\mathbf{\Phi}_{\text{task}, 1}, \dots , \mathbf{\Phi}_{\text{task}, \mathcal{S}}] \in \mathbb{R}^{ d \times \mathcal{S} }$ through matrix trace $\Tr(\mathbf{\Phi}_{\text{task}}\mathbf{\Omega}^{-1}\mathbf{\Phi}_{\text{task}}^{T})$. Recall that each client in our setting may only have a partial set of tasks in the labels of its training dataset, but still needs to make predictions for tasks it does not have in its labels during test time. This regularizer helps clients relate its own tasks to tasks in other clients.  Intuitively, it determines how closely two tasks $i$ and $j$ are related. The closer $\mathbf{\Phi}_{\text{task} , i}$ and $\mathbf{\Phi}_{\text{task} , j}$ are, the larger $\mathbf{\Omega}_{i,j}$ will be. If $\mathbf{\Omega}$ is an identity matrix, then each node is independent to each other. But as our results show, there is often a strong correlation between different molecular properties. This compels us to use a federated learning model.  

Figure~\ref{fig:framework2}-a depicts the FedGMTL framework where clients'  graph classifier  weights are using an FL server. While the above formulation enhances the FMTL with a constrained regularizer that can be used for GNN learning, we still need to solve the final challenge, which is to remove the reliance on a central server to perform the computations in  Eq. \eqref{eq:fedgmtl}.
Therefore, we propose a Decentralized Graph Multi-Task Learning framework, \texttt{SpreadGNN}, and utilize our novel optimizer, \textit{Decentralized Periodic Averaging SGD (DPA-SGD)} to extend the FMTL framework to our setting. The decentralized framework is shown in the Figure~\ref{fig:framework2}-c. Note that each client's learning task remains the same but the aggregation process differs in \texttt{SpreadGNN}.


\subsection{\texttt{SpreadGNN}: Serverless Federated MTL for GNNs}
\label{sec:de-gmtl}

\begin{figure}[h!]
  \includegraphics[width=0.65\linewidth]{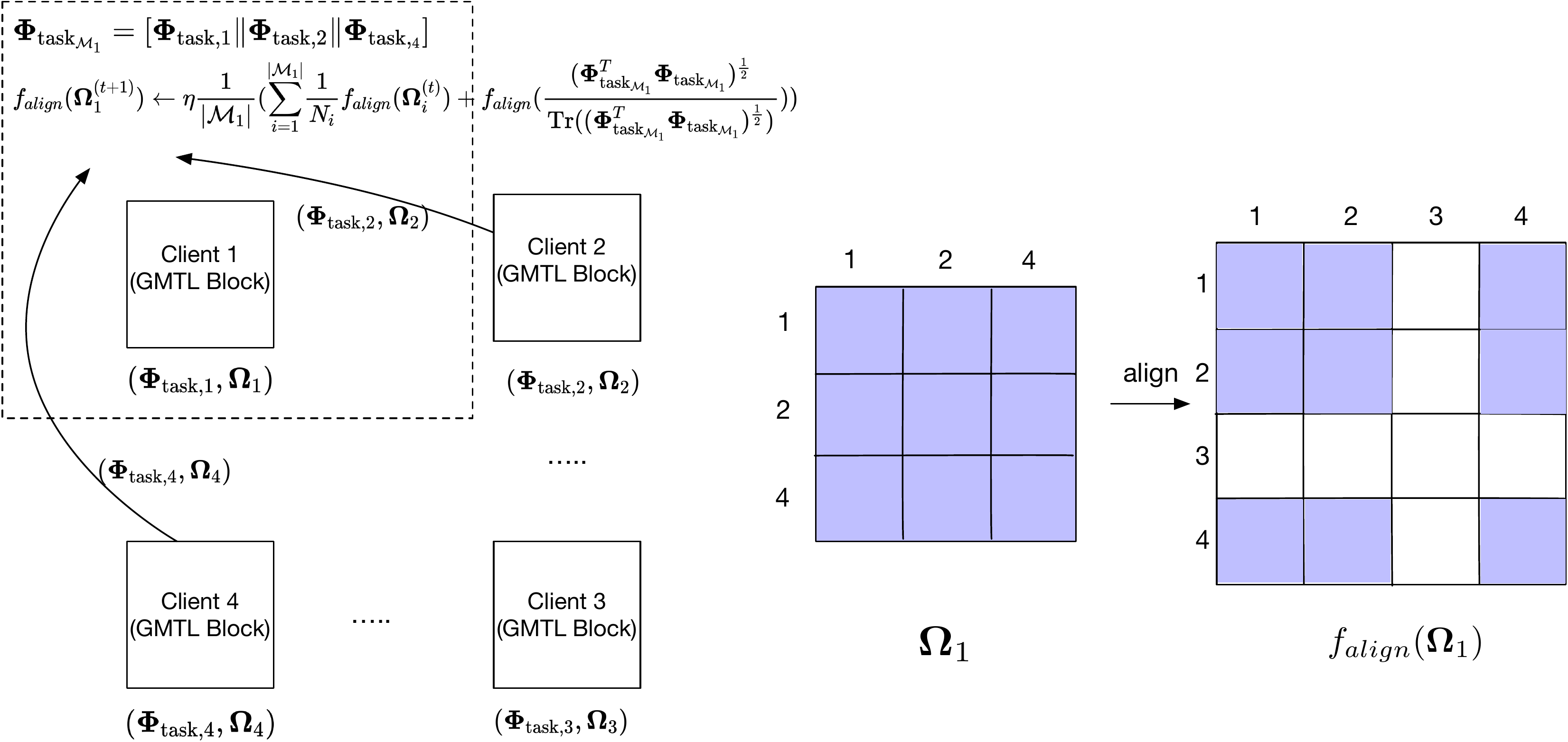}
  \centering
  \caption{Decentralized Multi-Task Learning Correlation Matrix Exchanging Algorithm}
  \label{fig:decentralized}
\end{figure}

To transition to the serverless setting, we introduce a novel optimization method called Decentralized Periodic Averaging SGD (DPA-SGD). The main idea of DPA-SGD is that each client applies SGD locally, and synchronizes all parameters with only its neighbors during a communication round that occurs every $\tau$ iterations. A decentralized system in which all clients are not necessarily connected to all other clients also makes it impossible to maintain one single task covariance matrix $\mathbf{\Omega}$. Thus we propose using distinct covariance matrices $\mathbf{\Omega}_k$ in each client that are efficiently updated using the exchange mechanism illustrated in Figure \ref{fig:decentralized}.  We formalize this idea as follows. Consider one particular client $m$ having task weights $ \mathbf{\Phi}_{\text{task},m} \in \mathcal{R}^{d \times \mathcal{S}_{m}}$ where $\mathcal{S}_{m}$ is the number of tasks that client $m$ has. Recall that the regularizer in Eq. \eqref{eq:eq6} is how various clients learn to predict tasks that are not within its $\mathcal{S}_{m}$.


In the decentralized setting, we emphasize that clients can collectively  learn the exhaustive set of tasks, even when clients may not have access to some of the classes in each multi-class label. That is and $\cup_{i} \mathcal{S}_{i}= \mathcal{S}$. Then, the new non-convex objective function can be defined as:


\begin{equation}
\begin{aligned}
&\begin{aligned}[t] \min_{\mathbf{\theta}, \mathbf{\Psi} , \mathbf{\Phi}_{\text{pool}} , \mathbf{\Phi}_{\text{task}} }  &\sum_{k=1}^{K}\frac{1}{N_k}\bigg[\bigg.\sum_{i=1}^{N_k}l(\hat{\mathbf{y}}_{i}^{(k)}(\boldsymbol{X}_{k},\boldsymbol{Z}_{k} ;\boldsymbol{W}_{k}),\mathbf{y}_{i}^{k}) + \frac{1}{2}\lambda_{1}\Tr(\mathbf{\Phi}_{\text{task}_{\mathcal{M}_{k}}}\mathbf{\Omega}_{k}^{-1}\mathbf{\Phi}_{\text{task}_{\mathcal{M}_{k}}}^{T})\bigg]\bigg. \\
&+ \frac{1}{2} \sum_{ \boldsymbol{\chi} \in \left\lbrace \mathbf{\theta}, \mathbf{\Psi} , \mathbf{\Phi}_{\text{pool}} , \mathbf{\Phi}_{\text{task}} \right\rbrace }\lambda_{\boldsymbol{\chi} } ||\boldsymbol{\chi}||_F^2, 
\end{aligned} \\
&\qquad \text{s.t.} \quad  \mathbf{\Omega}_k \ge 0 \qquad \text{and} \quad \Tr(\mathbf{\Omega}_k) = 1, \quad k = 1, 2, ... ,K.
\end{aligned}
\end{equation}
where $\boldsymbol{W}_{k} = \left\{\mathbf{\theta}, \mathbf{\Psi} , \mathbf{\Phi}_{pool} , \mathbf{\Phi}_{\text{task},k} \right\}$ is the set of all learnable weights for client $k$,  $\mathcal{M}_k =  {k} \cup \mathcal{N}_k$ is the neighbor set for client $k$ including itself. This gives rise to: $\mathbf{\Phi}_{\text{task}_{\mathcal{M}_k}} = [ \mathbf{\Phi}_{\text{task,1}} \| \mathbf{\Phi}_{\text{task,2}} \| \dots \| \mathbf{\Phi}_{\text{task},|\mathcal{M}_k|} ] \in \mathbb{R}^{d\times|\mathcal{S}_{\mathcal{M}_k}|}$ which is the task weight matrix for client $k$ and its neighbors and || represents the row-wise union operation. The matrix $\mathbf{\Omega}_k\in\mathbb{R}^{|\mathcal{S}_{\mathcal{M}_k}|\times{|\mathcal{S}_{\mathcal{M}_k}|}}$ represents the correlation amongst all the available tasks for the set $\mathcal{M}_k$. 

 To solve this non-convex problem, we apply the alternating optimization method presented in \cite{zhang_convex_2012}, where  $\boldsymbol{W}_{k}$ and $\mathbf\Omega_{k}$ are updated in an alternative fashion.\\
\textbf{Optimizing} $\boldsymbol{W}_{k}$ : For simplicity, we define $\underline{\boldsymbol{\Omega}} = \{ \boldsymbol{\Omega}_{1}, \boldsymbol{\Omega}_{2}, \dots ,\boldsymbol{\Omega}_{K} \}$ to represent the set of correlation matrices for all clients. Fixing $\underline{\boldsymbol{\Omega}}$, we can use SGD to update $\boldsymbol{W}_{k}$ jointly. Let $L =\sum_{k=1}^{K} \frac{1}{N_{k}} \sum_{i=1}^{N_{k}} l(\hat{y}_{i}^{(k)}(\boldsymbol{X}_{i}^{k},\boldsymbol{Z}_{i}^{k} \; \boldsymbol{W}_{k}),\mathbf{y}_{i}^k)$. Then, our problem can then be reformulated as:
\begin{equation}
 G(\boldsymbol{W}_{k} |\underline{\boldsymbol{\Omega}}) = L + \sum_{k=1}^{K} \frac{1}{N_{k}} \sum_{i=1}^{N_{k}} [  \frac{1}{2} \lambda_{1} \Tr(\mathbf{\Phi}_{\text{task}_{\mathcal{M}_{k}}} \mathbf{\Omega}_{k}^{-1} \mathbf{\Phi}_{\text{task}_{\mathcal{M}_{k}}} )] + \frac{1}{2} \sum_{\boldsymbol{\chi} \in \{\mathbf{\theta}, \mathbf{\Psi},\mathbf{\Phi}_{\text{pool}},\mathbf{\Phi_{\text{task}}}\}} \lambda_{\boldsymbol{\chi}}{\lVert\boldsymbol{\chi} \rVert}_{F}^{2}.
\end{equation}
 where the summation in (8) is amongst all the nodes connected to node $k$. Then the gradient formulations for each node are:
\begin{equation}
\frac{\partial{G(\boldsymbol{W}_{k} | \underline{\boldsymbol{\Omega}})}}{\partial{\mathbf{\Phi}_{\mathrm{task}_{\mathcal{M}_{k}}}}} = \frac{\partial{L}}{\partial{\mathbf{\Phi}_{\text{task}_{\mathcal{M}_{k}}}}} + \lambda_{1}\sum_{i=1}^{|\mathcal{M}_{k}|}\frac{1}{N_i}{\mathbf{\Phi}_{\text{task}_{\mathcal{M}_{k}}}}\mathbf\Omega_{i}^{-1}+ \lambda_{2}\mathbf{\Phi}_{\text{task}_{\mathcal{M}_{k}}}
\end{equation}
\begin{equation}
\frac{\partial{G(\boldsymbol{W}_{k} \| \underline{\boldsymbol{\Omega}})}}{\partial{\boldsymbol{\chi}}}=\frac{\partial{L}}{\partial{\boldsymbol{\chi}}} + \lambda_{\boldsymbol{\chi}}\boldsymbol{\chi}, \quad \forall \boldsymbol{\chi} \in \{ \mathbf{\theta}, \mathbf{\Psi} , \mathbf{\Phi}_{\text{pool}}\}\end{equation}
\\
\textbf{Optimizing} $\mathbf{\Omega_{k}} \in \underline{\boldsymbol{\Omega}}$: In \cite{zhang_convex_2012}, an analytical solution for  $\mathbf{\Omega}$ is equal to  \( \frac{(\mathbf{\Phi}_{\mathrm{task}}^{T} \mathbf{\Phi}_{\mathrm{task}})^{\frac{1}{2}}}{\Tr((\mathbf{\Phi}_{\mathrm{task}}^{T} \mathbf{\Phi}_{\mathrm{task}})^{\frac{1}{2}})}\).
 However, this solution is only applicable for  the centralized case. This is because The missing central node forbids  averaging parameters globally. So here we propose a novel way to update each $\mathbf{\Omega_{k}} \in \underline{\boldsymbol{\Omega}}$:
\begin{equation}
f_{align}(\mathbf\Omega_{k}^{(t+1)}) \leftarrow \eta\frac{1}{|\mathcal{M}_k|}({\sum_{i=1}^{|\mathcal{M}_k|}}\frac{1}{N_i} f_{align}(\mathbf\Omega_{i}^{(t)}) + f_{align}(\frac{(\mathbf{\Phi}_{\mathrm{task}_{\mathcal{M}_k}}^T \mathbf{\Phi}_{\mathrm{task}_{\mathcal{M}_k}})^\frac{1}{2}}{\Tr((\mathbf{\Phi}_{\mathrm{task}_{\mathcal{M}_k}}^T \mathbf{\Phi}_{\text{task}_{\mathcal{M}_k}})^{\frac{1}{2}})}))
\end{equation}
The first averaging term can incorporate the nearby nodes correlation into its own. It should be noticed that each $\mathbf{\Omega}_{i}$ may have a different dimension (different number of neighbors), so this averaging algorithm is based on the node-wised alignment in the global $\mathbf{\Omega}$. The $f_{align}$ function is illustrated in Figure \ref{fig:decentralized}. Here, $f_{align}$ operates on each client and   The second term captures the new correlation between its neighbors as shown in Figure \ref{fig:decentralized}. We refer readers to  to Algorithm \ref{alg:dpasgd-summary} in Appendix \ref{appendix:proof} 

\subsubsection{Convergence Properties of DPA-SGD}
In this section, we present our convergence analysis for DPA-SGD. Our analysis explains why DPA-SGD works well in the federated setting. In the previous section, we present DPA-SGD algorithm. Before any formalism, we first introduce a binary symmetric node connection matrix $\mathbf{M} \in \mathbb{R}^{K \times K}$ to record the connections in the network, where $\mathbf{M}_{i,j}$ is a binary value that indicates whether node $i$ and $j$ are connected or not. We assume that $\mathbf{M}$ satisfies the following: (a) $\mathbf{M} \mathbf{1}_{K} =  \mathbf{1}_{K}$ (b) $\mathbf{M}^{T} = \mathbf{M}$ (c) $ \max \{ |\lambda_{2}(\mathbf{M})|, \dots, |\lambda_{K}(\mathbf{M})| \} < \lambda_{1}(\mathbf{M}) = 1 $ 
where  $\lambda$ is the eigenvalues of $\mathbf{M}$. Note that, if $\mathbf{M}=\mathbf{I}$ (Identity matrix), then every nodes are independent and update the parameters respectively. If $\mathbf{M}=\mathbf{J}$ (one for each element), the model is degenerated into centralized model. Our convergence analysis satisfies the assumptions \cite{bottou_optimization_2018} listed in Appendix \ref{appendix:proof}. Here, we present mathematically simple and convenient  update rule of DPA-SGD:
\begin{equation}
\label{DHA-SGD-equation}
    \mathbf{X}_{t+1} = \left( \mathbf{X}_{t} - \eta \mathbf{G}_{t} \right) \cdot \mathbf{M}_{t},
\end{equation} where $\mathbf{X}_{t} = \left[ \mathbf{x}_{t}^{(1)},,, \mathbf{x}_{t}^{(K)} \right]$ is the matrix of our interest  (e.g. $\boldsymbol{W}_{k}^{(t)}$), $\mathbf{G}_{t}$ is the gradient  and $\mathbf{M}_{t}$ is the connection matrix at time t. When $t \mod \tau = 0, \mathbf{M}_{t} = \mathbf{M}$, otherwise, $\mathbf{M}_{t}$ equals to $\mathbf{I}_{K}$. In equation (\ref{DHA-SGD-equation}), multiplying $\mathbf{I}_{K}/K$ on both side, defining the averaged model $\mathbf{u}_{t} = \mathbf{X}_{t} \frac{\mathbf{I}_{K}}{K} $, we obtain the following update:
\begin{equation}
\label{u-define}
\begin{split}
    \mathbf{u}_{t+1} = \mathbf{u}_{t} - \eta \mathbf{g}_{t} = \mathbf{u}_{t} - \eta \left[ \frac{1}{K} \sum\limits_{i=1}^K g(\mathbf{x}_{t}^{(i)})   \right]
\end{split}
\end{equation}

Next, we will present our analysis on the convergence of the above-averaged model $\mathbf{u}_{t}$. For non-convex optimization, previous works on SGD convergence analysis use the \textit{average squared gradient norm} as an indicator of convergence to a stationary point [\citealp{bottou_optimization_2018}].

\begin{theorem}[\textbf{Convergence of DPA-SGD}]
 \label{DPA-SGD-theorem}
Assume that Assumption 1 and Definition (\ref{u-define}) hold. Then, if the learning rate $\eta$ satisfies the following condition:
 \begin{equation}
 \begin{split}
     \eta L + \frac{\eta^{2}L^{2}\tau^{2}}{1-\zeta} \left(\frac{2\zeta^{2}}{1+\zeta} + \frac{2\zeta}{1-\zeta} + \frac{\tau-1}{\tau} \right) \leq 1
 \end{split}
\end{equation}
 where $\tau$ is the averaging period(one synchronization per $\tau$ local updates), and $\zeta = \max \{ |\lambda_{2}(\mathbf{M})|, \dots,  |\lambda_{m}(\mathbf{M})|\}$, and all local
 models are initialized at a same point $\mathbf{x}_{0}$, then after K iterations the average squared gradient norm is bounded as
 \begin{equation}
 \label{t1}
     \mathbb{E} \Bigg[ \frac{1}{T} \sum\limits_{t=1}^T {\lVert\nabla F(\mathbf{u}_{t}) \rVert}^{2} \Bigg] \leq \frac{2[F(\mathbf{x}_{1})-F_{inf}]}{\eta T} + \frac{\eta L \sigma^{2}}{K} + 
       \eta^{2}L^{2}\sigma^{2} \left( \frac{1+\zeta^{2}}{1-\zeta^{2}}\tau - 1 \right)
 \end{equation}
 \end{theorem}

 Theorem \ref{DPA-SGD-theorem} shows that DPA-SGD algorithm is a trade-off between convergence rate and system optimization including communication-efficiency(communication rounds) and convergence speed.  find a recent work \cite{wang_cooperative_2018} has got the same result in an unified framework. Our work differs from this work in that it does not provide adequate theoretical analysis and empirical evaluation for federated learning. The proof of Theorem \ref{DPA-SGD-theorem} is given in the Appendix \ref{appendix:proof}.

%% file: sections/exps.tex
\section{Experiments}
\label{sec:experiments}

\subsection{Setup}

\paragraph{Implementation.} All experiments are conducted on a single GPU server equipped with 2 Nvidia Geforce GTX 1080Ti GPUs and an AMD Ryzen 7 3700X 8-Core Processor. Our models are built on top of the FedML framework \citep{chaoyanghe2020fedml} and PyTorch \citep{pytorch}.

\begin{table}[t!]
    \centering
    \captionof{table}{Dataset summary used in our experiments.  }
    \label{tab:datasets_summary}
    \setlength\tabcolsep{3pt}
    \begin{tabular}{c ||c c c c c c }
    \toprule
        \textbf{Dataset}  & \textbf{ \# Molecules} & \textbf{ Avg \# Nodes} & \textbf{ Avg \# Edges} & \textbf{\# Tasks} & \textbf{Task Type}  & \textbf{Evaluation Metric}  \\
         \midrule
         SIDER  & 1427 & 33.64 & 35.36 & 27 & Classification & ROC-AUC \\ 
         Tox21 & 7831 & 18.51 & 25.94 & 12  & Classification  & ROC-AUC \\
         MUV & 93087 & 24.23 & 76.80 & 17  & Classification  & ROC-AUC \\
         QM8 & 21786 & 7.77 & 23.95 & 12  & Regression  & MAE \\
    \bottomrule
    \end{tabular}
    \label{table:datasets}
\end{table}{}

\textbf{Multi-task Dataset.} We use molecular datasets from the MoleculeNet \citep{wu2018moleculenet} machine learning benchmark in our evaluation. In particular, we evaluate our approach on molecular property prediction datasets described in Table \ref{tab:datasets_summary}. The label for each molecule graph is a vector in which each element denotes a property of the molecule. Properties are binary in the case of classification or continuous values in the case of regression. As such, each multitask dataset can adequately evaluate our learning framework.

\textbf{Non-I.I.D. Partition for Quantity and Label Skew.} We introduce non-I.I.D.ness in two additive ways. The first is a non-I.I.D. split of the training data based on quantity. Here we use a Dirichlet distribution parameterized by $\alpha$ to split the training data between clients. Specifically, the number of training samples present in each client is non-I.I.D. The second source of non-I.I.D.ness is a label masking scheme designed to represent the scenario in which different clients may possess partial labels as shown in Figure \ref{fig:framework}. More specifically, we mask out a subset of classes in each label on every client. In our experiments, the sets of unmasked classes across all clients are mutually exclusive and collectively exhaustive. This setting simulates a worst case scenario where 
no two clients share the same task. However our framework is just as applicable when there is label overlap. Such masking, introduces a class imbalance between the clients making the label distribution non-I.I.D. as well.

\textbf{Models.} In order to demonstrate that our framework is agnostic to the choice of GNN model, we run experiments on two different GNN architectures: GraphSAGE \citep{graphsage} and GAT \citep{gat}.

\textbf{Network Topology.} We first evaluate our framework in a complete topology in which all clients are connected to all other clients to measure the efficacy of our proposed regularizer. We then perform ablation studies on the number of neighbors of each client to stress test our framework in the more constrained setting.

\textbf{Hyperparameters.} We use Adam \citep{adam} as the client optimizer in all of our experiments. For our Tox21, MUV and QM8 we use an 8 client topology. For SIDER we use a 4 client topology. A more comprehensive hyperparameter list for network topology and  models can be found in the Appendix \ref{appendix:hps}.

\subsection{Results}

We use a central, server dependent \texttt{FedAvg} system as the baseline of comparison. More specifically, all clients are involved in the averaging of model parameters in every averaging round. 

\begin{table}[t!]
    \centering
    \begin{tabular}{r| ccc | ccc }
    \toprule
    & \multicolumn{3}{c|}{\textbf{GraphSAGE}} & \multicolumn{3}{c}{\textbf{GAT}} \\
    \cline{2-7}
       & {\small \textbf{FedAvg}} & {\small \textbf{FedGMTL}} & {\small \textbf{SpreadGNN}} & {\small \textbf{FedAvg}} & {\small \textbf{FedGMTL}} & {\small \textbf{SpreadGNN}} \\
    \hline
    \textbf{SIDER}  &  0.582   & 0.629 & \textbf{0.5873} & 0.5857 & 0.61 & \textbf{0.6034} \\
    \textbf{Tox21}   &  0.5548   &  0.6664 & \textbf{0.585} & 0.6035 & 0.6594 & \textbf{0.6056} \\
    \textbf{MUV} & 0.6578 & 0.6856 & \textbf{0.703} & 0.709 & 0.6899 & \textbf{0.713} \\
    \textbf{QM8} & 0.02982 & 0.03624 & \textbf{0.02824} & 0.0392 & 0.0488 & \textbf{0.0315} \\
    \bottomrule
    \end{tabular}{}
    \caption{Results on the molecular property prediction task. SpreadGNN uses a complete topology: all clients communicated with each other. Communication period $\tau = 1$}
    \label{tab:results}
\end{table}

\begin{figure}[t!]
    \begin{subfigure}{.5\textwidth}
    \centering
    \includegraphics[scale = 0.5]{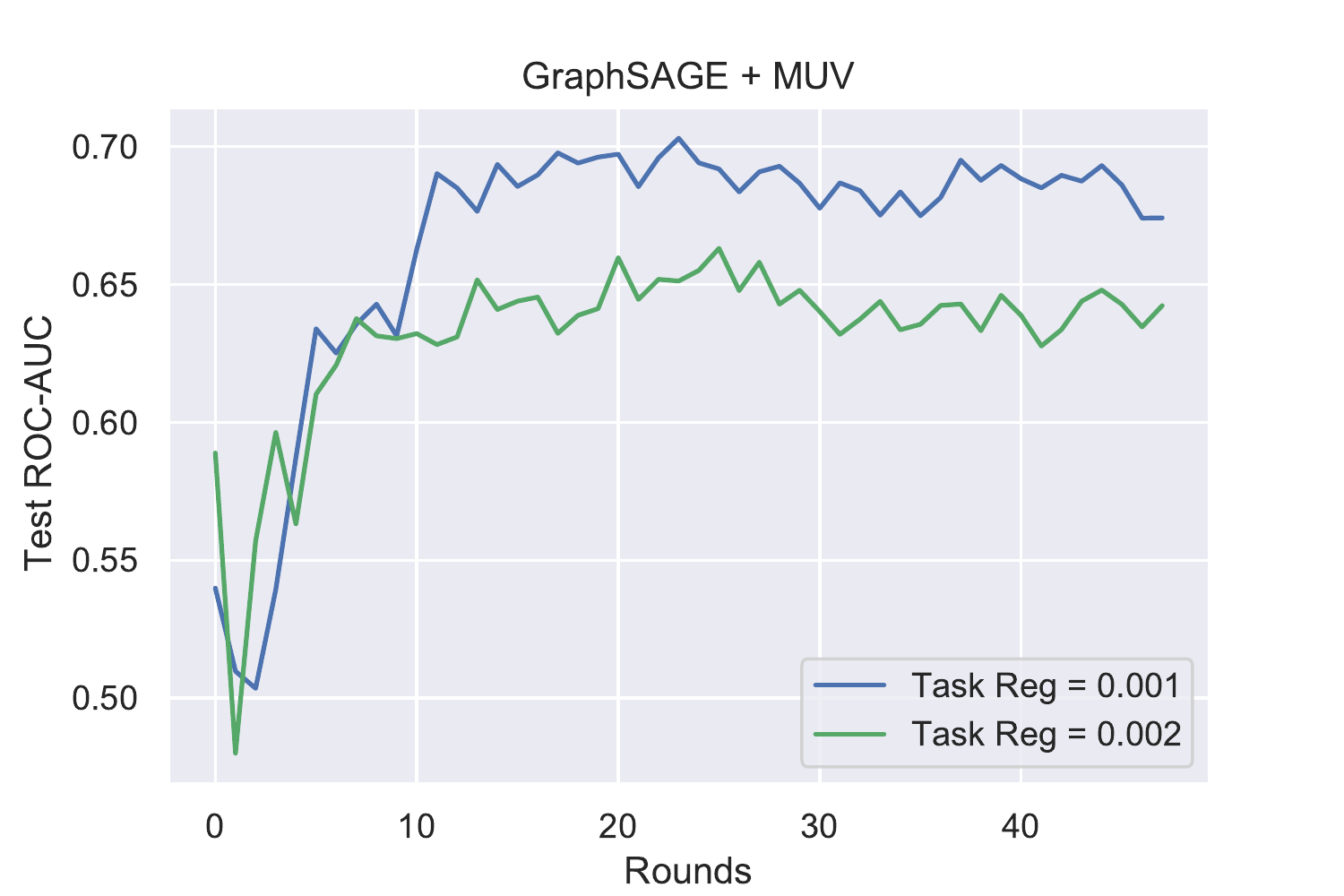}
    \end{subfigure}%
    \begin{subfigure}{.5\textwidth}
    \centering
    \includegraphics[scale = 0.5]{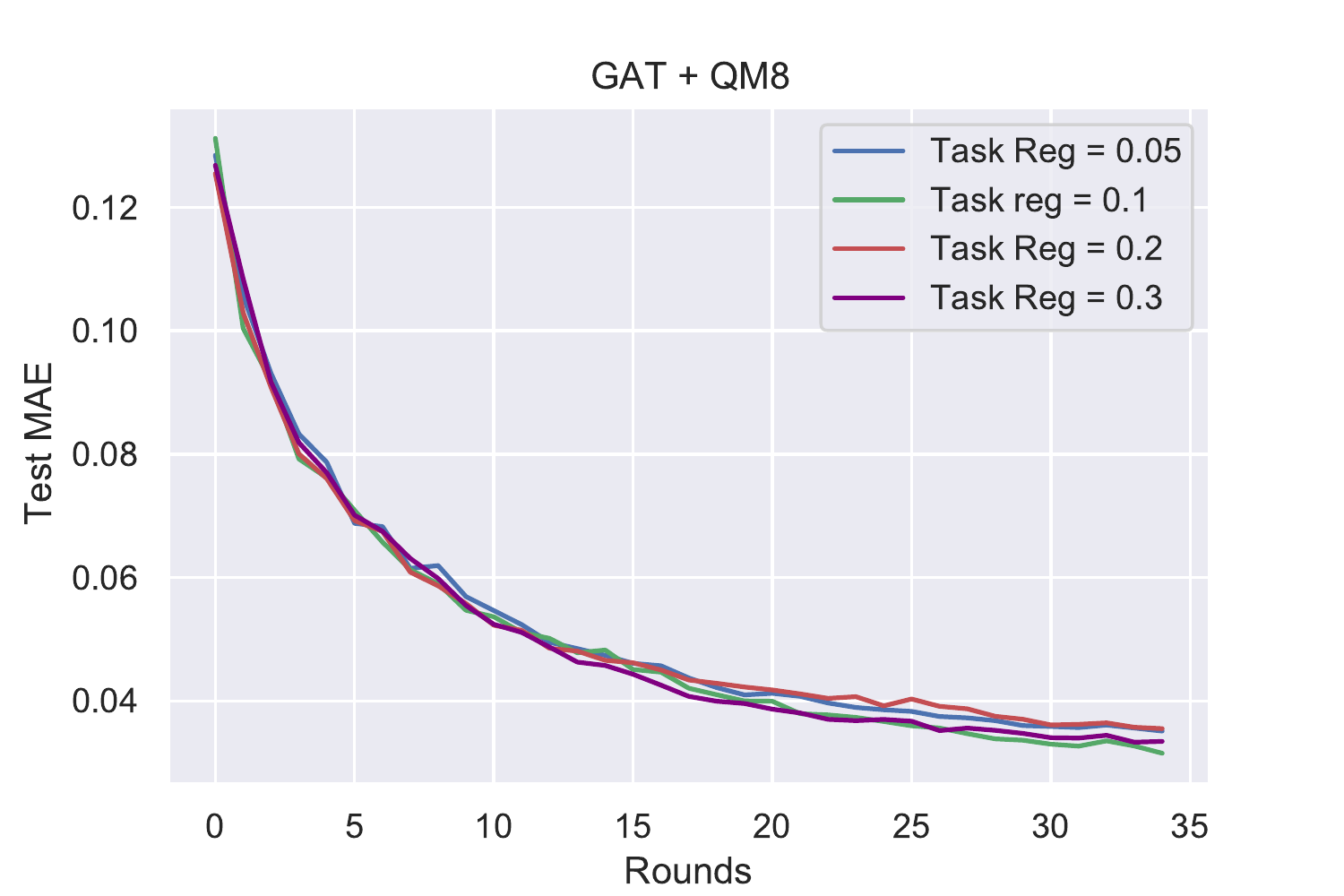}
    \end{subfigure}
    \caption{Effect of Task-Relationship Regularizer on Learning}
    \label{fig:effectoftaskreg}
\end{figure}

\begin{figure}[t!]
    \centering
    \includegraphics[scale = 0.5]{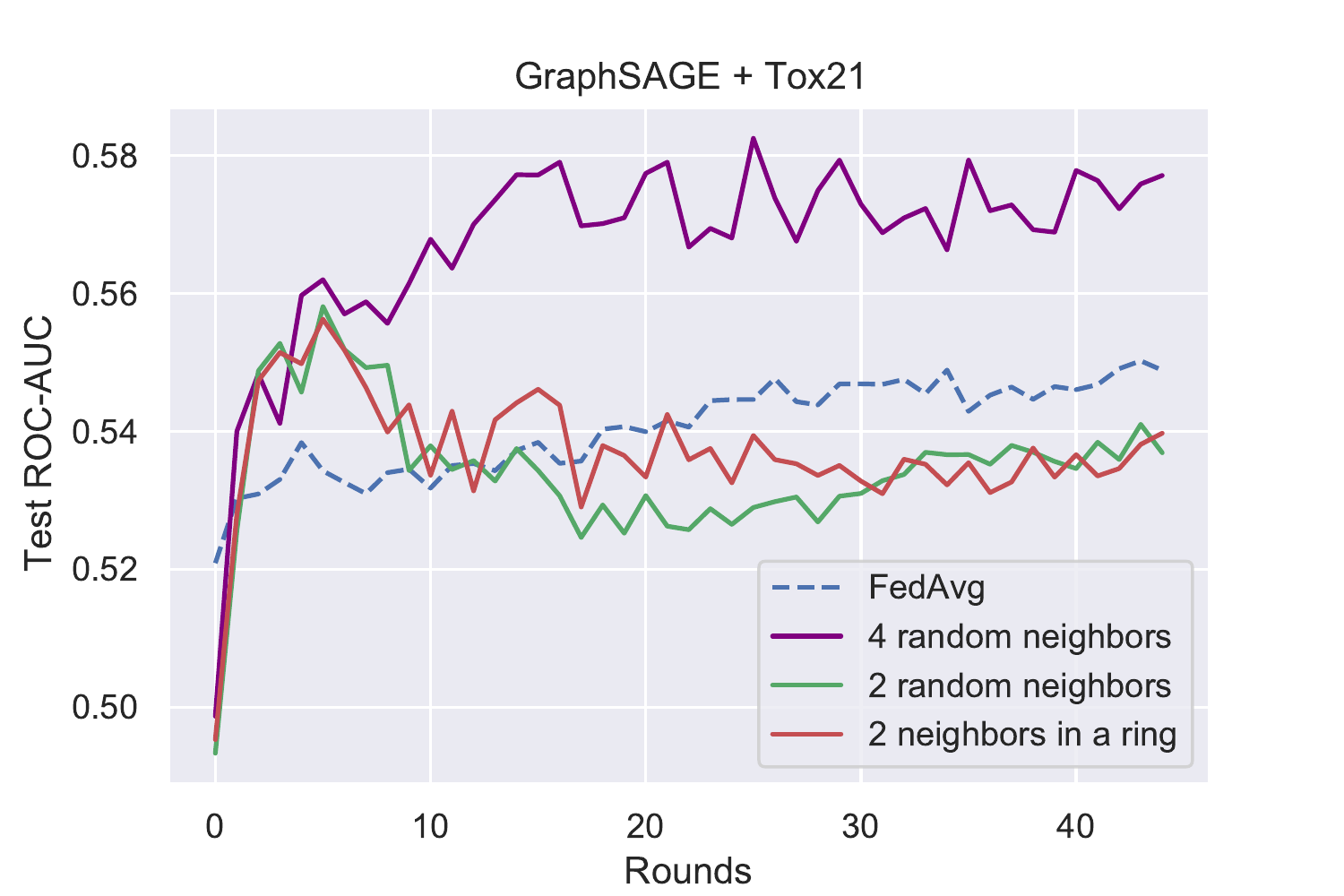}
    \caption{Effect of Topology on Graphsage + Tox21}
    \label{fig:ablation1}
\end{figure}

Our results summarized in Table \ref{tab:results} demonstrate that \texttt{SpreadGNN} (third column) outperforms a centralized federated learning system that uses FedAvg (first column) when all clients can communicate with all other clients. This shows that by using the combination of the task regularizer in equation \ref{eq:eq6} and the DPA-SGD algorithm, we can eliminate the dependence on a central server and enable clients to learn more effectively in the presence of missing molecular properties in their respective labels. Additionally, the results also show that our framework is agnostic to the type of GNN model being used in the molecular property prediction task since both GraphSAGE and GAT benefit from our framework. Our framework also works in the case a trusted central server is available (second column). The presence of a trusted central server improves accuracy in a few scenarios. However, SpreadGNN provides competing performance in a more realistc setting where central server is not feasible.    

\subsection{Sensitivity Analysis}
\label{sec:hps}

\textbf{Task Regularizer.} Figure \ref{fig:effectoftaskreg} illustrates the effect of $\lambda_1$ on both regression as well as classification tasks. Interestingly, regression is much more robust to variation in $\lambda_1$ while classification demands more careful tuning of $\lambda_1$ to achieve optimal model performance. This implies that the different properties in the regression task are more independent than the properties in the classification task. 

\textbf{Network Topology.} The network topology dictates how many neighbors each client can communicate with in a communication round. While Table \ref{tab:results} shows that SpreadGNN outperforms FedAvg in a complete topology, Figure \ref{fig:ablation1} shows that our framework performs outperforms FedAvg even when clients are constrained to communicate with fewer neighbors. We can also see that it's not just $n_{neighbors}$ that matters, the topology in which clients are connected does too. When $n_{neighbors} = 2$, a ring topology outperforms a random topology, as a ring guarantees a path from any client to any other client. Thus, learning is shared indirectly between all clients. The same is not true in a random topology. More ablation studies on the the network topology are included in the Appendix \ref{appendix:more-exps}. 

\textbf{Period.} The communication period $\tau$ is another important hyperparameter in our framework. In general, our experiments suggest that a lower period is better, but this is not always the case. We include an ablation study on $\tau$ to support this claim in the Appendix \ref{appendix:more-exps}. 






%% file: sections/related.tex
\section{Related Works}

\textbf{Molecular Representation Learning.} 
 \cite{rogers2010extended} encode the neighbors of atoms in the molecule into a fix-length vector to obtain vectors-space representations. To improve the expressive power of chemical fingerprints,  \cite{duvenaud2015convolutional, coley2017convolutional} use CNNs to learn rich molecule embeddings for downstream tasks like property prediction. \cite{kearnes2016molecular,schutt2017quantum,schutt2017schnet} explore the graph convolutional network to encode molecular graphs into neural fingerprints. To better capture the interactions among atoms, \cite{gilmer2017neural} proposes to use a message passing framework and \cite{yang2019analyzing,klicpera2020directional} extend this framework to model bond interactions. 

\textbf{FL}. Early examples of research into federated learning include \cite{konecny_federated_2015, konecny_federated_2016, mcmahan_communication-efficient_2016}. To address both statistical and system challenges, \cite{smith_federated_2017} and \cite{caldas_federated_2018} propose a multi-task learning framework for federated learning and its related optimization algorithm, which extends early works in distributed machine learning \cite{shalev-shwartz_stochastic_2013, yang_trading_2013, yang_analysis_2013,jaggi_communication-efficient_2014, ma_adding_2015, smith_cocoa:_2016}. The main limitation of \cite{smith_federated_2017} and \cite{caldas_federated_2018}, however, is that strong duality is only guaranteed when the objective function is convex, which can not be generalized to the non-convex setting.\cite{ahmed_scalable_2014, jin_collaborating_2015, mateos-nunez_distributed_2015, wang_distributed_2016, baytas_asynchronous_2016, liu_distributed_2017} extends federated multi-task learning to the distributed multi-task learnings setting, but  not only limitation remains same , but also nodes performing same amount of work is prohibitive in FL. 

\textbf{Federated Graph Neural Networks}. \citep{suzumura2019towards} and \citep{mei2019sgnn} use computed graph statistics for information exchange and aggregation to avoid node information leakage. \citep{jiang2020federated, zhou2020privacy} utilize cryptographic approaches to  incorporate into GNN learning. \citep{wang2020graphfl} propose a hybrid of federated and meta learning to solve the semi-supervised graph node classification problem in decentralized social network datasets. \citep{meng2021crossnode} uses an edge-cloud partitioned GNN model for spatio-temporal traffic forecasting tasks over sensor networks. The previous works do not consider graph learning in a decentralized setting.

\textbf{Stochastic Gradient Descent Optimization}. In large scale distributed machine learning problems learning, synchronized mini-batch SGD is a well-known method to address the communication bottleneck by increasing the computation to communication ratio \cite{dean_large_nodate, li_scaling_2014, cui_exploiting_2014}. It is shown that FedAvg \cite{konecny_federated_2016} is a special case of local SGD which allow nodes to perform local updates and infrequent synchronization between them to communicate less while converging fast \cite{stich_local_2018, wang_cooperative_2018, yu_parallel_2018, lin_dont_2018}. Decentralized SGD, another approach to reducing communication, was successfully applied to deep learning [\citealp{jin_how_2016, jiang_collaborative_2017, lian_can_2017}]. Asynchronous SGD is a potential method that can alleviate synchronization delays in distributed learning [\citealp{recht_hogwild:_2011, cui_exploiting_2014, gupta_model_2016, mitliagkas_asynchrony_2016, dutta_slow_2018}], but existing asynchronous SGD doesn't fit for federated learning because the staleness problem is particularly severe due to the reason of heterogeneity in the federated setting \cite{dai_toward_2018}.


%% file: sections/conclusion.tex
\section{Conclusion}
\label{sec:conclusion}

In this work, we propose \texttt{SpreadGNN}, a framework to train graph neural networks in a serverless federated system. We motivate our framework through a realistic setting, in which clients involved in molecular machine learning research cannot share data with each other due to privacy and competition. Moreover, we further recognize that clients may only possess partial labels of their data. Through experimental analysis, we show for the first time, that training graph neural networks in a federated manner does not require a centralized topology and that our framework can address the non-iidness in dataset size and label distribution across clients. \texttt{SpreadGNN} can outperform a central server depend baseline even when clients can only communicate with a few neighbors. To support our empirical results, we also provide a convergence analysis of the \textit{DPA-SGD} optimization algorithm used by \texttt{SpreadGNN}.


\section*{Broader impact}
\label{sec:broader_impact}
Our method can be used to train graph neural networks for molecular property prediction tasks in a serverless manner. Our method can have immense societal benefits such as accelerated drug discovery. One limitation of our method is that although federated learning can protect privacy to a certain extent, recent studies have shown that it cannot guarantee that data or models are not leaked. Therefore, if researchers want to deploy our method to the actual production environment as a public service, we recommend analyzing inherent security and privacy risks comprehensively and adequately. It is necessary to add security and privacy-related components, such as Differential Privacy and Secure Aggregation.

%% file: sections/appendix-dataset_details.tex
\section{Algorithm Sketch}

\begin{algorithm}[h]
  \caption{ \texttt{SpreadGNN} : Serverless Multi-task Federated Learning for Graph Neural Networks }
  \begin{algorithmic} [1]
	 \REQUIRE initial parameters for each node $\boldsymbol{W}_{k}^{(t=0)} = \{\mathbf{\theta}^{(t=0)}, \mathbf{\Psi}^{(t=0)} , \mathbf{\Phi}_{\text{pool}}^{(t=0)} , \mathbf{\Phi}_{\text{task},k}^{(t=0)} \}$ and $\underline{\boldsymbol{\Omega}}^{(t=0)} = (\Omega_{1}^{(t=0)}, \Omega_{2}^{(t=0)},...,\Omega_{K}^{(t=0)})$; learning rate $\eta$; maximum number of global iterations $T$, maximum number of client epochs $E$; communication period $\tau$.
	 \FOR {all nodes: $k=1,2,...,K$ \textbf{in parallel}}
	 \FOR {$t = 1$ to $T$ do}
	 \FOR {$k=1$ to $E$(epoch loop) do}
	 \FOR {$m \in MB$ (mini-batch loop) do}
	 \STATE Read a minibatch $m$ \\
	 \STATE Calculate gradient: $g(\boldsymbol{W}_{k}^{t,m} ) =  \partial{G(\boldsymbol{W}_{k}^{t,m} |\underline{\boldsymbol{\Omega}}^{t,m}) }  / \partial{\boldsymbol{W}_{k}^{t,m} }$
	 \STATE Update the local $k^{th}$ optimization variables:\\
	 $\mathbf{\boldsymbol{W}}_{k}^{(t+1,m)} \gets \boldsymbol{W}_{k}^{(t,m)}-\eta g(\boldsymbol{W}_{k}^{(t,m)})$\\
	 $\Omega_k^{(t+1,m)} \gets (\mathbf{\Phi}_{\mathrm{task}_{\mathcal{M}_k}}^T \mathbf{\Phi}_{\mathrm{task}_{\mathcal{M}_k}})^\frac{1}{2} /\Tr((\mathbf{\Phi}_{\mathrm{task}_{\mathcal{M}_k}}^T \mathbf{\Phi}_{\text{task}_{\mathcal{M}_k}})^{\frac{1}{2}})$
	 \ENDFOR
	 \ENDFOR
	 \IF{$t \mod \tau = 0$}
	 \STATE Perform aggregation and alignment over neighbors for node $k$: \\
	 $\boldsymbol{W}_{k}^{(t+1)} \gets (\sum_{j=1}^{|\mathcal{M}_k|}\frac{1}{N_j} \boldsymbol{W}_{k}^{(t)}) / |\mathcal{M}_k|$,\\
	 $f_{align}(\Omega_k^{t+1}) \gets \eta ( \sum_{j= \mathcal{M}_k  \backslash k } \frac{1}{N_j} f_{align}(\Omega_j^{(t)}) +f_{align}(\frac{(\mathbf{\Phi}_{\mathrm{task}_{\mathcal{M}_k}}^T \mathbf{\Phi}_{\mathrm{task}_{\mathcal{M}_k}})^\frac{1}{2}}{\Tr((\mathbf{\Phi}_{\mathrm{task}_{\mathcal{M}_k}}^T \mathbf{\Phi}_{\text{task}_{\mathcal{M}_k}})^{\frac{1}{2}})}))/|\mathcal{M}_k|$
	 \ENDIF
	 \ENDFOR
	 \ENDFOR
  \end{algorithmic}
  \label{alg:dpasgd-summary}
\end{algorithm}

\section{Dataset Details}
\label{appendix:dataset}

Table \ref{tab:datasets_summary} summarizes the necessary information of benchmark datasets \citep{wu2018moleculenet}. The details of each dataset are listed below:

\begin{itemize}
    \item \texttt{SIDER} \citep{kuhn2016sider}, or Side Effect Resource, the dataset consists of marketed drugs with their adverse drug reactions. 
    \item \texttt{Tox21}\citep{tox21} is a dataset which records  the toxicity of compounds. 
    \item \texttt{MUV} \citep{rohrer2009maximum} is a subset of PubChem BioAssay processed via  refined nearest neighbor analysis. Contains  17  tasks for around 90 thousand compounds and is specifically designed for validation of virtual screening techniques.

    \item \texttt{QM8} \cite{Ramakrishnan_2015} is composed from a  recent study on modeling quantum mechanical calculations of electronic spectra and excited state energy of small molecules.
\end{itemize}

\subsection{Feature Extraction Procedure for Molecules}
\label{appendix:exp:feature}

The feature extraction is in two steps:
\begin{enumerate}
    \item  Atom-level feature extraction and Molecule object construction using RDKit \citep{landrum2006rdkit}.
    \item Constructing graphs from molecule objects using NetworkX \citep{SciPyProceedings_11}.
\end{enumerate}
Atom features, shown in  Table \ref{tab:atomfea}, are the atom features we used. It's exactly the same set of features as used in \citep{grover}. 
\begin{table}[htbp]
\caption{Atom features}
\begin{center}
\begin{tabular}{lll}
\toprule
\multicolumn{1}{c}{\bf Features}  &\multicolumn{1}{c}{\bf Size} &\multicolumn{1}{c}{\bf Description}
\\ \midrule 

atom type & 100  &  Representation of atom (e.g., C, N, O), by its atomic number\\
    formal charge  & 5  & An integer electronic charge assigned to atom  \\
    number of bonds & 6 & Number of bonds the atom is involved in \\
    chirality & 5 & Number of bonded hydrogen atoms\\
    number of H & 5 & Number of bonded hydrogen atoms\\
    atomic mass & 1 & Mass of the atom, divided by 100\\
    aromaticity & 1 & Whether this atom is part of an aromatic system\\
    hybridization & 5 &  SP, SP2, SP3, SP3D, or SP3D2\\
    \bottomrule
\end{tabular}

\label{tab:atomfea}
\end{center}

\end{table}

%% file: sections/appendix-hparams.tex
\section{Model Hyperparameters}
\label{appendix:hps}

\subsection{Model Architecture}

As explained in section \ref{sec:formulation} our model is made up of a GNN and Readout. The GNNs we use are GAT \cite{gat} and GraphSAGE \cite{graphsage}. Each accepts input node features $X_v \in \mathcal{R}^{|V_M
| \times d_{input}}$ and outputs node embeddings $h_v \in \mathcal{R}^{|V_M
| \times d_{node}}$, $v \in V_M$. Where $V_M$ is the set of atoms in molecule $M$. Given the output node embeddings from the GNN the Readout function we use is defined as follows:

\begin{equation*}    \boldsymbol{R}_{\mathbf{\Phi_{\text{pool}}}, \mathbf{\Phi_{\text{task}}}}\left(h_v, X_v \right) = \text{MEAN}\left(\mathbf{ReLU}\left(\mathbf{\Phi_{\text{task}}}\left(\mathbf{ReLU}\left(\mathbf{\Phi_{\text{pool}}}\left(X_v \| h_v\right)\right)\right)\right)\right)
\end{equation*}

where $\|$ represents the row wise concatenation operation.  $\mathbf{\Phi_{\text{pool}}} \in \mathcal{R}^{(d_{node} + d_{input}) \times d_{pool}}$ and $\mathbf{\Phi_{\text{task}}} \in \mathcal{R}^{d_{pool} \times d_{out}}$ are learnable transformation matrices. $d_{out}$ represents the number of classes/tasks present in the classification label. The $\text{MEAN}$ operation here is a column wise mean. Note that while our general description of the readout in section \ref{sec:formulation} does not include the input features as part of the input, we find that including the input features leads to better generalization.

\subsection{Hyperparameter Configurations}

For each task, we utilize grid search to find the best results. Table~\ref{tab:hyper-parameters-fed} lists all the hyper-parameters ranges used in our experiments. All hyper-parameter tuning is run on a single  GPU. The best hyperparameters for each dataset and model are listed in Table \ref{tab:results-with-hps}. The batch-size is kept 1. This pertains to processing a single molecule at a time. The number of GNN layers were fixed to 2 because having too many GNN layers result in over-smoothing phenomenon as shown in \citep{li2018deeper}. For all experiments, we used Adam optimizer.

\begin{table}[!ht]
  \centering
  \caption{Hyperparameter Range for Experiments}
  \resizebox{1\textwidth}{!}{
    \begin{tabular}{l|l|l}
    \toprule
    hyper-parameter & Description & Range \\
    \midrule
    Learning rate & Rate of speed at which the model learns.  & $\left[0.00015, 0.001, 0.0015, 0.0025, 0.015, 0.15 \right]$ \\
    Dropout rate & Dropout ratio & $\left[ 0.3, 0.5, 0.6 \right]$ \\
    Node embedding dimension ($d_{node}$) & Dimensionality of the node embedding & 64 \\
    Hidden layer dimension & GNN hidden layer dimensionality &  64 \\
    Readout embedding dimension ($d_{pool}$) & Readout Hidden Layer Dimensionality &  64 \\
    Graph embedding dimension ($d_{out}$) & Dimensionality of the final graph embedding &  64 \\
    Attention heads & Number of attention heads required for GAT & 1-7 \\
    Alpha & LeakyRELU parameter used in GAT model & 0.2  \\
    Rounds & Number of federating learning rounds & 150 \\
    Epoch  & Epoch of  clients & 1 \\
    Number of clients & Number of users in a federated learning round & 4-10 \\
    Communication Period  & Exchanging Period between clients & 1 \\
    \bottomrule
    \end{tabular}
}
  \label{tab:hyper-parameters-fed}
\end{table}%

\begin{table}[t!]
  
    \caption{Hyperparameters used in our experiments. For SpreadGNN we use a communication period $\tau = 1$ and a complete topology (all clients connected to all other clients) in all experiments.}
{ 
  \centering
  \resizebox{.95\textwidth}{!}{
    \begin{tabular}{|r |c| ccc | ccc| }
    \toprule
    &  \multicolumn{1}{|c|}{}
    & \multicolumn{3}{c|}{ {\small \textbf{GraphSAGE}} } & \multicolumn{3}{c|}{{\small \textbf{GAT}}} \\
    \cline{3-8}
       &
        \multicolumn{1}{c|}{\small \textbf{Parameters}}  & {\small \textbf{FedAvg}} & {\small \textbf{FedGMTL}} & {\small \textbf{SpreadGNN}} & {\small \textbf{FedAvg}} & {\small \textbf{FedGMTL}} & {\small \textbf{SpreadGNN}} \\
    \hline
    \multirow{7}{*}{{\small SIDER }}
    & {\footnotesize \textbf{ROC-AUC Score} } & 0.582 & \textbf{0.629} & 0.5873 & 0.5857 & \textbf{0.61} & 0.603 \\\cline{2-8}
    & Partition alpha & 0.2 & 0.2 & 0.2 & 0.2 & 0.2 & 0.2 \\\cline{2-8}
    & Learning rate  & 0.0015 & 0.0015 & 0.0015 & 0.0015 & 0.0015 & 0.0015 \\\cline{2-8}
    & Dropout rate  & 0.3 & 0.3 & 0.3 & 0.3 & 0.3 & 0.3 \\\cline{2-8}
    & Node embedding dimension  & 64 & 64 & 64 & 64 & 64 & 64  \\\cline{2-8}
    & Hidden layer dimension  & 64 & 64 & 64 & 64 & 64 & 64  \\\cline{2-8}
    & Readout embedding dimension & 64 & 64 & 64 & 64 & 64 & 64  \\\cline{2-8}
    & Graph embedding dimension & 64 & 64 & 64 & 64 & 64 & 64  \\\cline{2-8}
    & Attention Heads & NA & NA & NA & 2 & 2 & 2 \\\cline{2-8}
    & Leaky ReLU alpha & NA & NA & NA & 0.2 & 0.2 & 0.2 \\\cline{2-8}
    & Number of Clients & 4 & 4 & 4 & 4 & 4 & 4 \\\cline{2-8}
    & Task Regularizer & NA & 0.001 & 0.001 & NA & 0.001 & 0.001 \\\cline{2-8}
    \hline
    \multirow{7}{*}{{\small Tox21 }}
    & {\footnotesize \textbf{ROC-AUC Score} } & 0.5548 & \textbf{0.6644} & 0.585 & 0.6035 & \textbf{0.6594} & 0.6056 \\\cline{2-8}
    & Partition alpha & 0.1 & 0.1 & 0.1 & 0.1 & 0.1 & 0.1  \\\cline{2-8}
    & Learning rate  & 0.0015 & 0.0015 & 0.0015  & 0.0015 & 0.0015 & 0.0015 \\\cline{2-8}
    & Dropout rate  & 0.3 & 0.3 & 0.3 & 0.3 & 0.3 & 0.3  \\\cline{2-8}
    & Node embedding dimension  & 64 & 64 & 64 &  64 & 64 & 64 \\\cline{2-8}
    & Hidden layer dimension  & 64 & 64 & 64 &  64 & 64 & 64 \\\cline{2-8}
    & Readout embedding dimension & 64 & 64 & 64 &  64 & 64 & 64  \\\cline{2-8}
    & Graph embedding dimension & 64 & 64 & 64 & 64 & 64 & 64  \\\cline{2-8}
    & Attention Heads & NA & NA & NA & 2 & 2 & 2 \\\cline{2-8}
    & Leaky ReLU alpha & NA & NA & NA & 0.2 & 0.2 & 0.2 \\\cline{2-8}
    & Number of Clients & 8 & 8 & 8 & 8 & 8 & 8 \\\cline{2-8}
    & Task Regularizer & NA & 0.001 & 0.001 & NA & 0.001 & 0.001 \\\cline{2-8}
    \hline
    \multirow{7}{*}{{\small MUV }}
    & {\footnotesize \textbf{ROC-AUC Score} } & 0.6578 & 0.6856 & \textbf{0.703} & 0.709 & 0.6899 & \textbf{0.713} \\\cline{2-8}
    & Partition alpha & 0.3 & 0.3 & 0.3 & 0.3 & 0.3 & 0.3 \\\cline{2-8}
    & Learning rate  & 0.001 & 0.001 & 0.001 & 0.0025 & 0.0025 & 0.0025 \\\cline{2-8}
    & Dropout rate  & 0.3 & 0.3 & 0.3 & 0.3 & 0.3 & 0.3 \\\cline{2-8}
    & Node embedding dimension  & 64 & 64 & 64 & 64 & 64 & 64 \\\cline{2-8}
    & Hidden layer dimension  & 64 & 64 & 64 & 64 & 64 & 64 \\\cline{2-8}
    & Readout embedding dimension & 64 & 64 & 64 & 64 & 64 & 64 \\\cline{2-8}
    & Graph embedding dimension & 64 & 64 & 64 & 64 & 64 & 64 \\\cline{2-8}
    & Attention Heads & NA & NA & NA & 2 & 2 & 2 \\\cline{2-8}
    & Leaky ReLU alpha & NA & NA & NA & 0.2 & 0.2 & 0.2 \\\cline{2-8}
    & Number of Clients & 8 & 8 & 8 & 8 & 8 & 8\\\cline{2-8}
    & Task Regularizer & NA & 0.001 & 0.001 & NA & 0.002 & 0.002 \\\cline{2-8}
    \hline
    \multirow{7}{*}{{\small QM8 }}
    & {\footnotesize \textbf{RMSE Score} } & 0.02982 & 0.03624 & \textbf{0.02824} & 0.0392 & 0.0488 & \textbf{0.0333} \\\cline{2-8}
    & Partition alpha & 0.5 & 0.5 & 0.5 & 0.5 & 0.5 & 0.5 \\\cline{2-8}
    & Learning rate  & 0.0015 & 0.0015 & 0.0015 & 0.0015 & 0.0015 & 0.0015 \\\cline{2-8}
    & Dropout rate  & 0.3 & 0.3 & 0.3 & 0.3 & 0.3 & 0.3 \\\cline{2-8}
    & Node embedding dimension  & 64 & 64 & 64 & 64 & 64 & 64 \\\cline{2-8}
    & Hidden layer dimension  & 64 & 64 & 64 & 64 & 64 & 64 \\\cline{2-8}
    & Readout embedding dimension & 64 & 64 & 64 & 64 & 64 & 64 \\\cline{2-8}
    & Graph embedding dimension & 64 & 64 & 64 & 64 & 64 & 64 \\\cline{2-8}
    & Attention Heads & NA & NA & NA & 2 & 2 & 2 \\\cline{2-8}
    & Leaky ReLU alpha & NA & NA & NA & 0.2 & 0.2 & 0.2 \\\cline{2-8}
    & Number of Clients & 8 & 8 & 8 & 8 & 8 & 8 \\\cline{2-8}
    & Task Regularizer & NA & 0.3 & 0.3 & NA & 0.3 & 0.3  \\\cline{2-8}

    \bottomrule
    \end{tabular}
}
}

    \label{tab:results-with-hps}
\end{table}

%% file: sections/appendix-ablation.tex
\section{Detailed Ablation Studies}
\label{appendix:more-exps}

\subsection{Effect of Communication Period \texorpdfstring{$\tau$}{Lg}}

\begin{figure}[ht!]
    \begin{subfigure}{.5\textwidth}
    \centering
    \includegraphics[scale = 0.5]{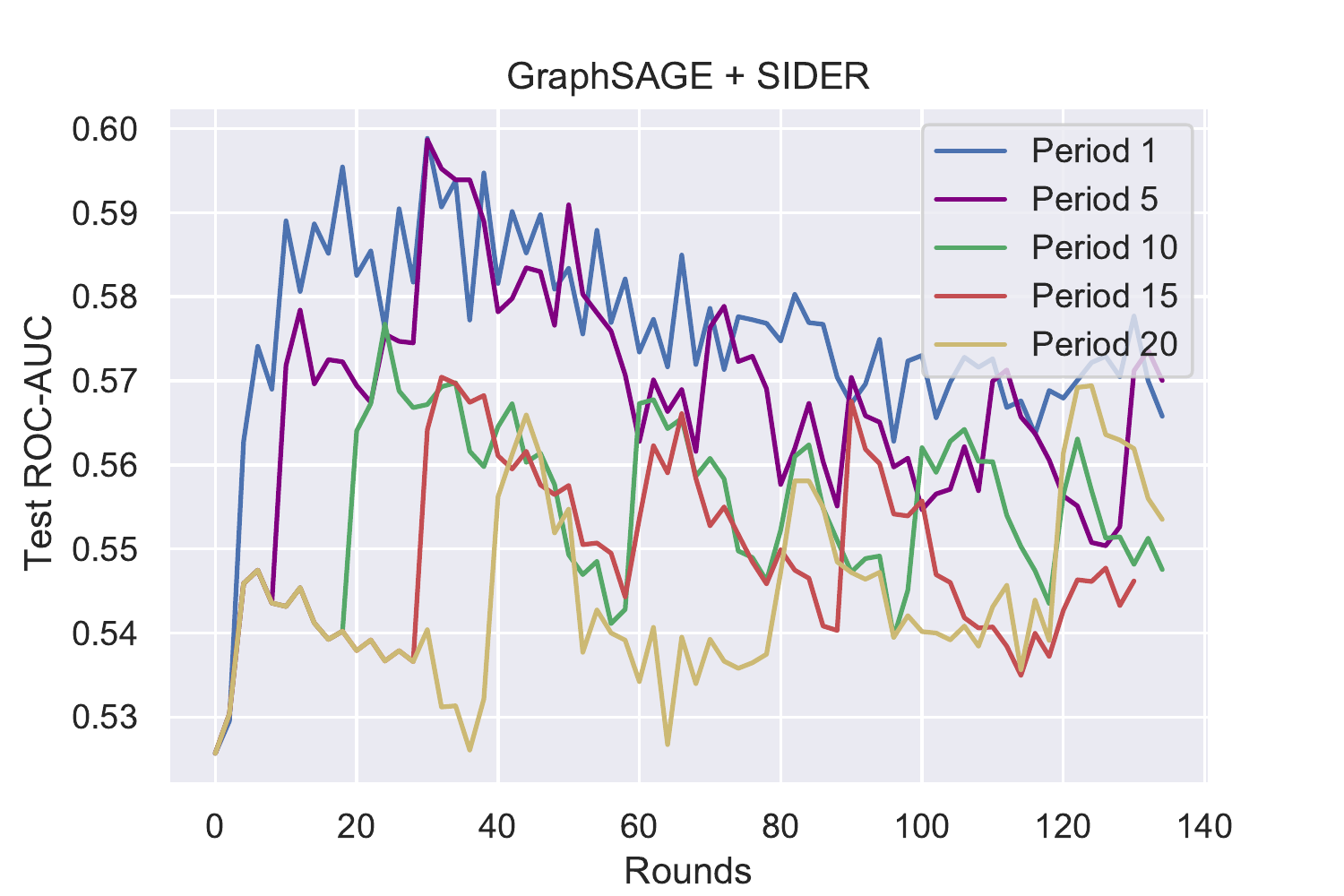}
    \end{subfigure}%
    \begin{subfigure}{.5\textwidth}
    \centering
    \includegraphics[scale = 0.5]{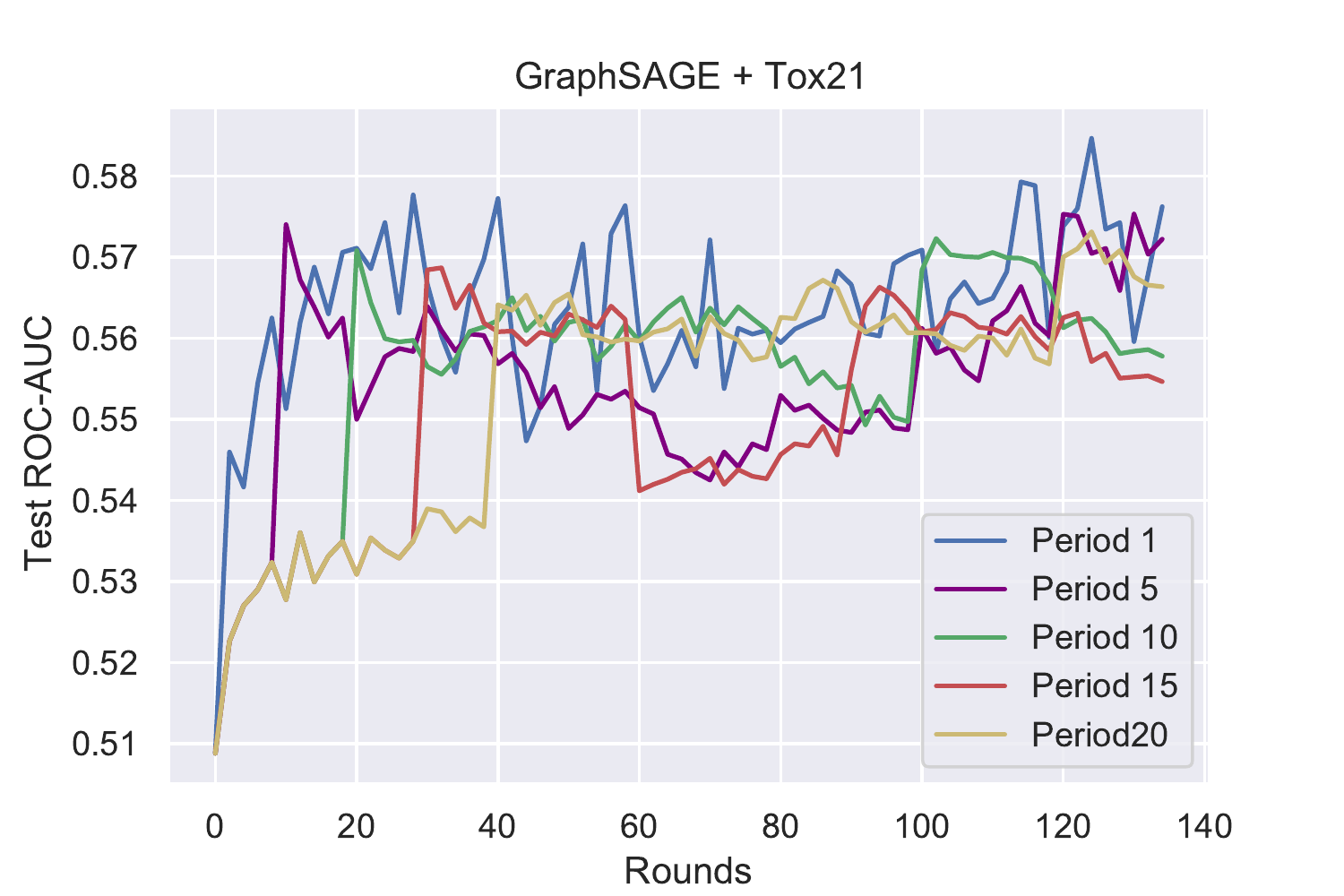}
    \end{subfigure}
    \caption{Effect of Communication Period $\tau$ on GraphSAGE Model}
    \label{fig:effectofperiod_sage}
\end{figure}

\begin{figure}[ht!]
    \begin{subfigure}{.5\textwidth}
    \centering
    \includegraphics[scale = 0.5]{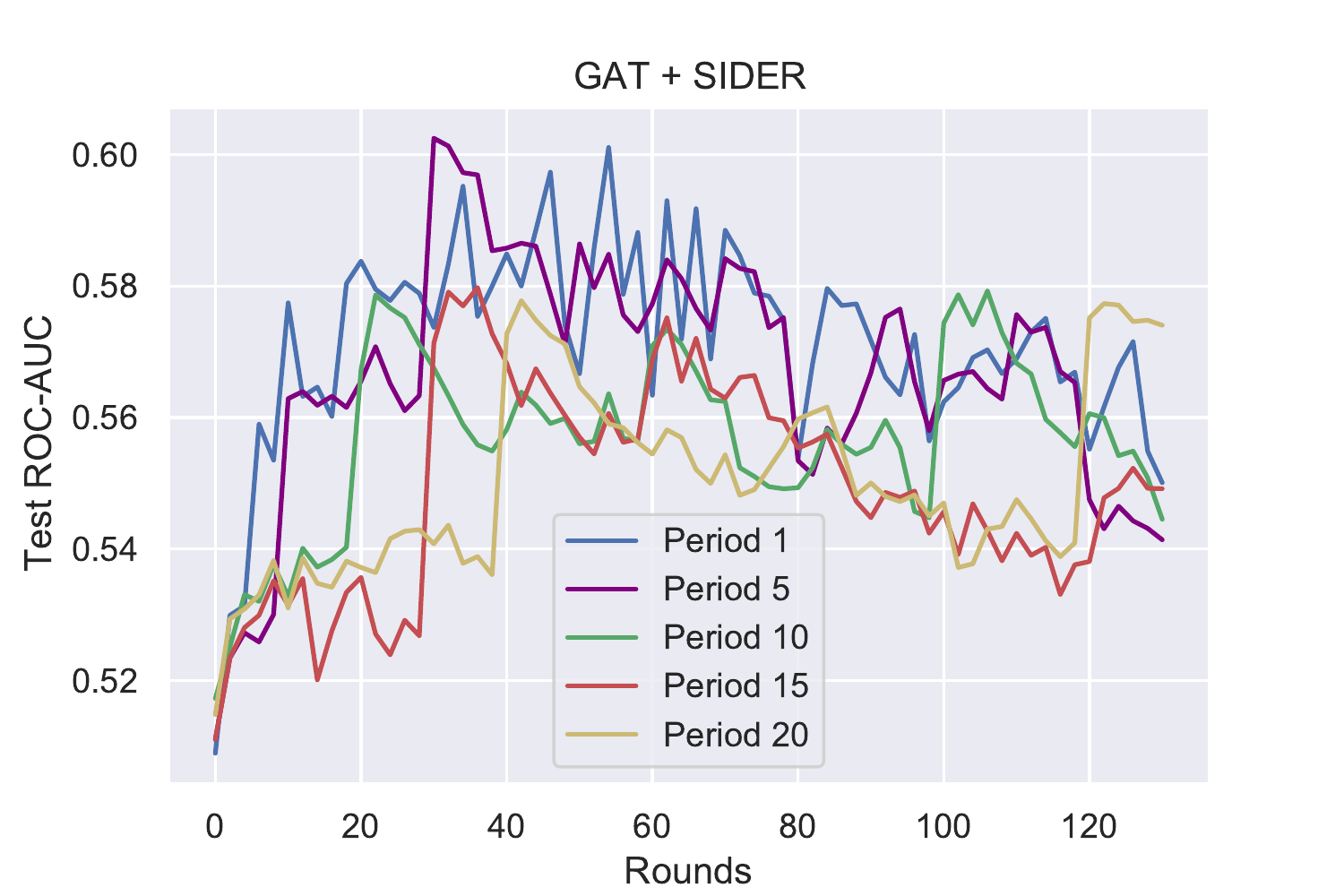}
    \end{subfigure}%
    \begin{subfigure}{.5\textwidth}
    \centering
    \includegraphics[scale = 0.5]{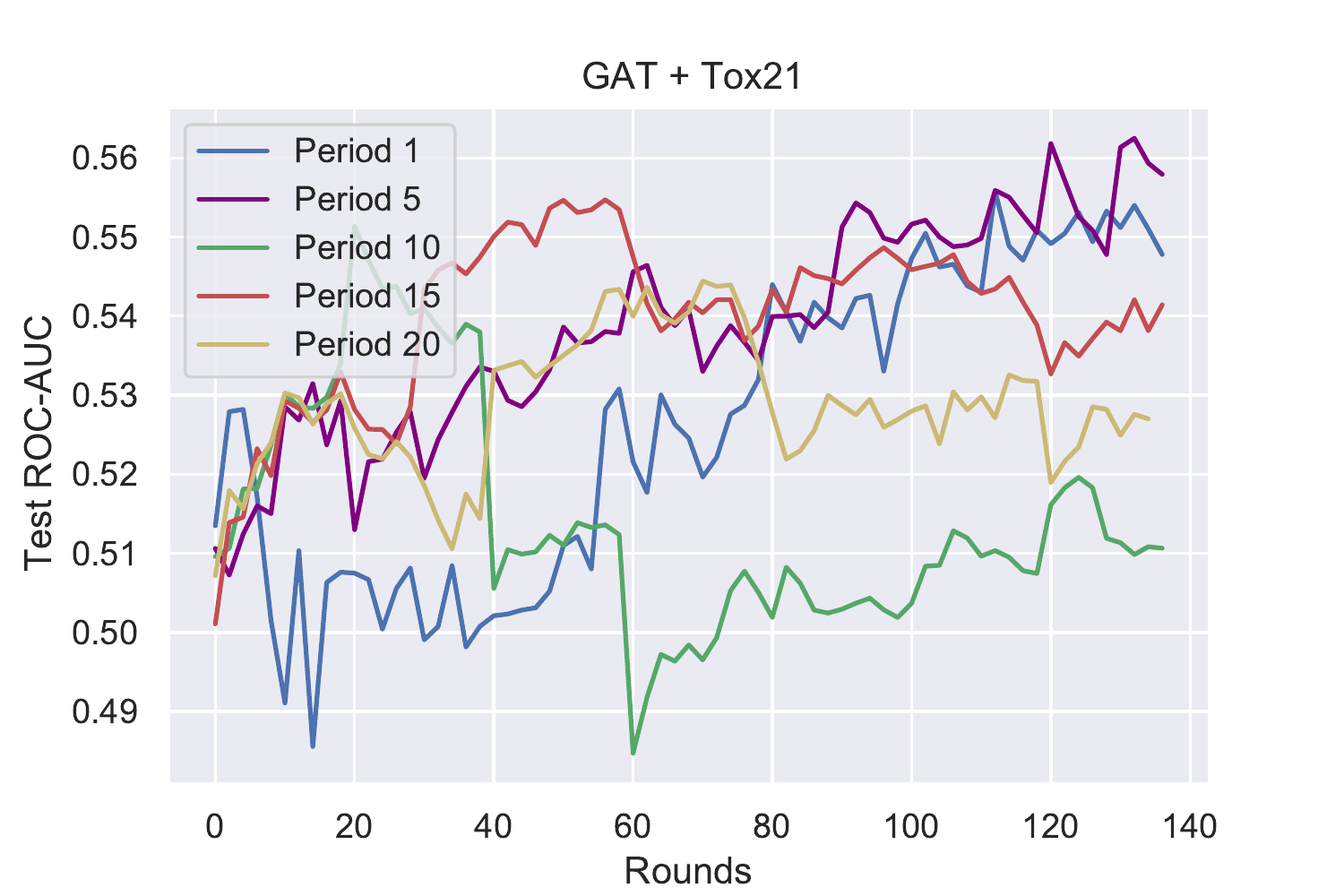}
    \end{subfigure}
    \caption{Effect of Communication Period $\tau$ on GAT Model}
    \label{fig:effectofperiod_gat}
\end{figure}

Figures \ref{fig:effectofperiod_sage} \& \ref{fig:effectofperiod_gat} illustrate the effect of communication period SIDER and Tox21 datasets. As we increase the communication period $\tau$ more, model performance decreases. However, selecting $\tau=5$ can sometimes be better than averageing \& exchanging each round. This indicates that, tuning $\tau$ is important for  while controlling the tradeoff between the performance and the running time.

\subsection{Effect of Serverless Network Topology}

\begin{figure}[ht!]
    \centering
    \includegraphics[scale=0.5]{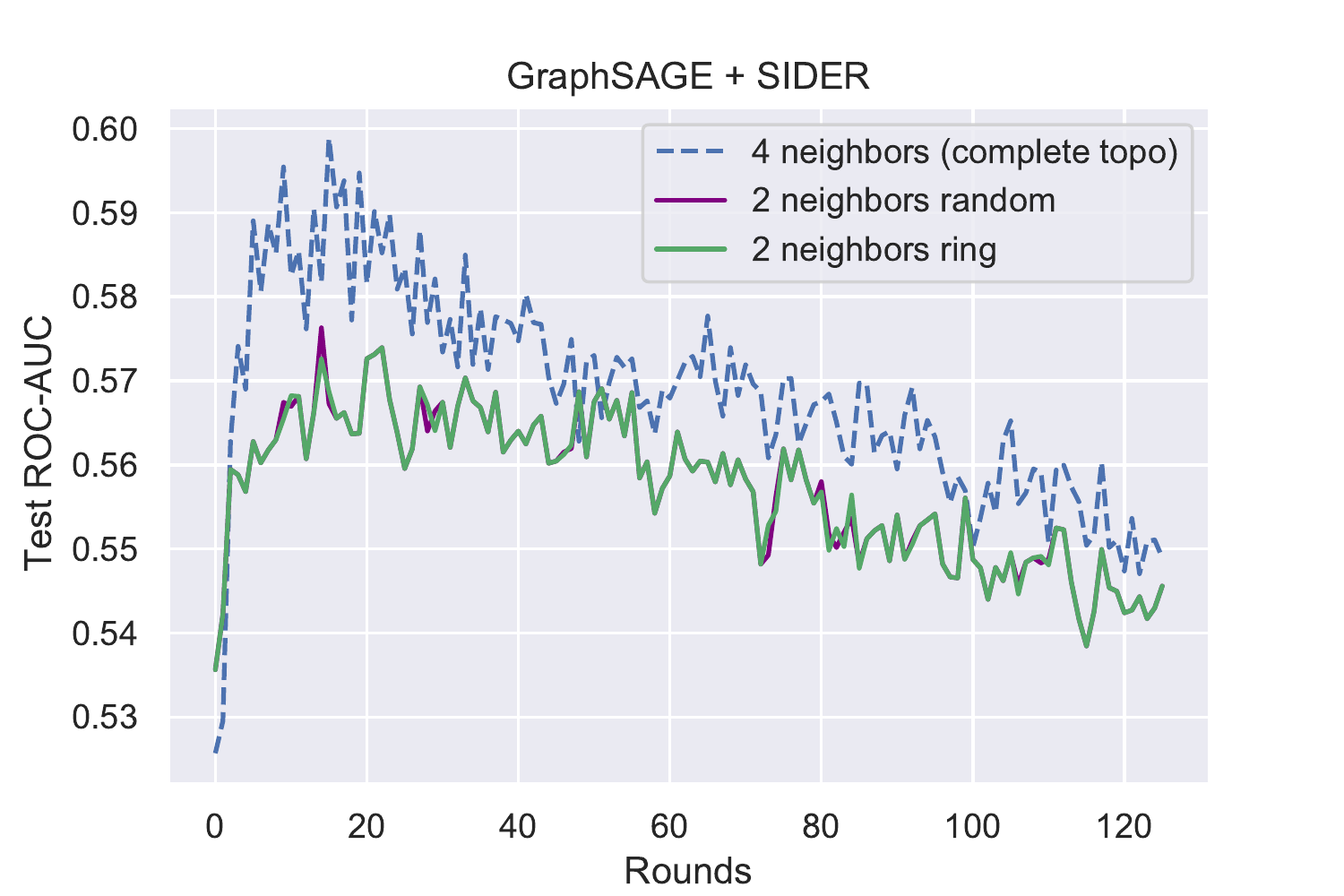}
    \caption{Effect of Topology on Graphsage + Sider using a 4 client network}
    \label{fig:moretopoablation}
\end{figure}

Figure \ref{fig:moretopoablation} illustrates the effect of varying topologies on SpreadGNN on the Sider dataset when using Graphsage as the GNN. The qualitative behavior is similar to Figure \ref{fig:ablation1}, in that when each client is connected to more neighbors, the local model on each client is more robust. However, when the total number of clients involved in the network is smaller, the effect of topology is understated and the total number of neighbors matters more. Recall that in an 8 client network, when each client was restricted to being connected to only 2 neighbors, random connections performed worse than a ring topology, meaning that the topology mattered as much as mere number of neighbors. However, in the case of the 4 client network in Figure \ref{fig:moretopoablation} there is a minimal difference between a 2 neighbor random configuration and a 2 neighbor ring configuration.  

%% file: sections/appendix-proof.tex
\section{ Proof for Convergence of DPA-SGD}

\label{appendix:proof}

In our analysis, we assume that following properties hold \cite{bottou_optimization_2018}:

\begin{itemize}
    \item The objective function $F(\cdot)$ is \textbf{L-Lipschitz}.
    \item $F(\cdot)$ is \textbf{lower bounded} by $F_{inf}$ such that $F(\cdot) \geq F_{inf}$.
    \item The full gradient of the objective function $F(\cdot)$ is approximated by stochastic gradient $g$ on a mini-batch $\epsilon_{i}$ with \textbf{an unbiased estimate}: $ \mathop{\mathbb{E}}_{\epsilon_{k}} [ g(\mathbf{x})] = \nabla F(\mathbf{x})$.
    \item The variance of stochastic gradient $g(\mathbf{x})$ given a mini-batch $\epsilon_{k}$ is \textbf{upper bounded} and  $\text{Var}_{\epsilon_{k}}(g(\mathbf{x})) \leq  \beta {\lVert \nabla F(\mathbf{x}) \rVert}^{2} + \sigma^{2}, \quad \exists \beta, \sigma^{2}\geq 0, \quad \forall k$.
\end{itemize}

In order to get a more clear view of our algorithm, we reformulate the loss function on each worker as follows:
\begin{align*}
f_k\left(\mathbf{\Gamma},\mathbf{\Phi}_{\text{task}_{\mathcal{M}_{k}}},\mathbf{\Omega}^{(k)};\mathbf{\xi}_{i}^{(k)}\right):=&\, \mathcal{L}\left( \hat{\mathbf{y}}_{i}^{(k)}(\boldsymbol{X}_{i,k},\boldsymbol{Z}_{i,k} ;\boldsymbol{W}_{k}),\mathbf{y}_i^{(k)}\right)+ \frac{1}{2}\lambda_1 \Tr(\mathbf{\Phi}_{\text{task}_{\mathcal{M}_{k}}} \mathbf{\Omega}_{k}^{-1} \mathbf{\Phi}_{\text{task}_{\mathcal{M}_{k}}}) \\
&+ \frac{1}{2} \sum_{ \boldsymbol{\chi} \in \left\lbrace   \mathbf{\theta}, \mathbf{\Psi} , \mathbf{\Phi}_{\text{pool}} , \mathbf{\Phi}_{\text{task}} \right\rbrace }\lambda_{\boldsymbol{\chi} } ||\boldsymbol{\chi}||_F^2,
\end{align*}
where we include all the parameters that is different on each worker to be $\mathbf{\delta}^{(k)}:=\{\mathbf{\Phi}_{\text{task}_{\mathcal{M}_{k}}}\}$,  $\mathbf{\Gamma}$ denotes the shared parameters, and $\xi_{i}^{(k)}$ is the random variable that denotes the data samples $(\boldsymbol{X}_{i,k},\boldsymbol{Z}_{i,k},\mathbf{y}_i^{(k)})$.

Therefore, the original objective function an be cast into the following form:
\begin{align*}
    F\left({\mathbf{\Gamma}},\mathbf{\delta}^{(1:K)},\mathbf{\Omega}^{(1:K)}\right) = \frac{1}{n}\sum_{i=1}^{K}\mathbb{E}_{\xi_{i}^{(k)}}f_k\left({\mathbf{\Gamma}},\mathbf{\delta}^{(k)},\mathbf{\Omega}^{(k)};\xi_{i}^{(k)}\right).
\end{align*}{}
Notice that the updating rule for ${\mathbf{\Gamma}}$, $\mathbf{\delta}^{(1:K)}$, and $\mathbf{\Omega}^{(1:K)}$ are different in that:
\begin{align*}
    \mathbf{\Gamma}_{t+1} =& \mathbf{\Gamma}_t - \eta\frac{\partial{F}}{\partial{\mathbf{\Gamma}_t}}\mathbf{M},\\
    \mathbf{\delta}^{(k)}_{t+1} =& \mathbf{\delta}_t^{(k)} - \eta \frac{\partial{F}}{\partial{\mathbf{\delta}_t^{(k)}}},\\
    \mathbf{\Omega}^{(k)}_{t+1} =& \text{argmin}_{\mathbf{\Omega}^{(k)}} F\left({\mathbf{\Gamma}},\mathbf{\delta}^{(1:K)}_t,\underline{\mathbf{\Omega}}^{(1:K)}_t\right).
\end{align*}{}
From the update rule of $\mathbf{\Omega}^{(1:K)}_t$, we know that
{\footnotesize 
\begin{align*}
    &\mathbb{E}_{t}F\left({\overline{\mathbf{\Gamma}}}_{t+1},\mathbf{\delta}^{(1:K)}_{t+1},\mathbf{\Omega}^{(1:K)}_{t+1}\right) - \mathbb{E}_{t}F\left({\overline{\mathbf{\Gamma}}}_{t},\mathbf{\delta}^{(1:K)}_{t},\mathbf{\Omega}^{(1:K)}_{t}\right) \\
    \leq & \mathbb{E}_{t}\left\langle \frac{\partial{F}}{\partial{\overline{\mathbf{\Gamma}}_t}}, \overline{\mathbf{\Gamma}}_{t+1} - \overline{\mathbf{\Gamma}}_t \right\rangle + \mathbb{E}_{t}\left\langle \frac{\partial{F}}{\partial{\mathbf{\delta}^{(1:K)}_{t}}}, \mathbf{\delta}^{(1:K)}_{t+1} - \mathbf{\delta}^{(1:K)}_{t} \right\rangle + \frac{L}{2}\mathbb{E}_{t}\left( \left\|\overline{\mathbf{\Gamma}}_{t+1} - \overline{\mathbf{\Gamma}}_t \right\|^2_F + \left\|\mathbf{\delta}^{(1:K)}_{t+1} - \mathbf{\delta}^{(1:K)}_{t} \right\|^2_F \right)\\
    \leq & -\eta\left( 1 - \frac{L\eta}{2}\right)  \mathbb{E}_{t}\left\|\frac{\partial{F}}{\partial{\mathbf{\delta}^{(1:K)}_{t}}}\right\|^2_F  - \frac{\eta}{2}\mathbb{E}_{t}\left\|\frac{\partial{F}}{\partial{\overline{\mathbf{\Gamma}}_t}}\right\|^2_F - \frac{\eta\left( 1 - L\eta\right)}{2}\mathbb{E}_{t}\left\|\frac{\partial{F}}{\partial{\mathbf{\Gamma}}_t}\frac{1_n}{n}\right\|^2_F + \frac{\eta}{2}\left\|\frac{\partial{F}}{\partial{\overline{\mathbf{\Gamma}}_t}} - \frac{\partial{F}}{\partial{\mathbf{\Gamma}}_t}\frac{1_n}{n}\right\|^2_F \\
& + \frac{L\eta^{2}(\sigma_{\Gamma}^{2} + \sigma_{\delta}^{2})}{2K}, \numberthis \label{final:eq1}
\end{align*}
}
 where the overline $\overline{*}$ denotes the expectation operation $\mathbb{E}$, $\sigma^2_{\Gamma}$ and $\sigma^2_{\delta}$ are the variance bounds for the stochastic gradients of $\mathbf{\Gamma}_t $ and $ \delta_t$ respectively. To estimate the upper bound for
$\mathbb{E}\left\|\frac{\partial{F}}{\partial{\overline{\mathbf{\Gamma}}_t}} - \frac{\partial{F}}{\partial{\boldsymbol{\Gamma}}_t}\frac{1_n}{n}\right\|^2_F$, we utilize convexity and Lipschitzness as follows:
\begin{align*}
\mathbb{E}\left\|\frac{\partial{F}}{\partial{\overline{\mathbf{\Gamma}}_t}} - \frac{\partial{F}}{\partial{\mathbf{\Gamma}}_t}\frac{1_n}{n}\right\|^2_F = & \frac{1}{n^2}\mathbb{E}{\left\|
	\sum_{i=1}^n\left(\frac{\partial{F_i}}{\partial{\overline{\mathbf{\Gamma}}_t}}  - \frac{\partial{F_i}}{\partial{\mathbf{\Gamma}}_t^{(i)}}\right)\right\|^2_F}\\
\leq & \frac{1}{n}\sum_{i=1}^n\mathbb{E}\left\|\frac{\partial{F_i}}{\partial{\overline{\mathbf{\Gamma}}_t}}  - \frac{\partial{F_i}}{\partial{\mathbf{\Gamma}}_t^{(i)}}\right\|^2_F\\
\leq & \frac{L^2}{n}\mathbb{E}\sum_{i=1}^n\left\|\overline{\mathbf{\Gamma}}_t-\mathbf{\Gamma}_t^{(i)}\right\|^2_F.
\end{align*}
Therefore, we re-write the lower bound \eqref{final:eq1} as
{\footnotesize
\begin{equation*}
-\eta\left( 1 - \frac{L\eta}{2}\right)  \mathbb{E}_{t}\left\|\frac{\partial{F}}{\partial{\mathbf{\delta}^{(1:K)}_{t}}}\right\|^2_F  - \frac{\eta}{2}\mathbb{E}_{t}\left\|\frac{\partial{F}}{\partial{\overline{\mathbf{\Gamma}}_t}}\right\|^2_F - \frac{\eta\left( 1 - L\eta\right)}{2}\mathbb{E}_{t}\left\|\frac{\partial{F}}{\partial{\mathbf{\Gamma}}_t}\frac{1_n}{n}\right\|^2_F + \frac{\eta L^2}{2n}\mathbb{E}_{t}\sum_{i=1}^n\left\|\overline{\mathbf{\Gamma}}_t-\mathbf{\Gamma}_t^{(i)}\right\|^2_F+ \frac{L\eta^2(\sigma_{\Gamma}^2 + \sigma_{\delta}^2)}{2K}. 
\end{equation*}
}

Summing the inequality above for all time-steps $t=0, \dots , T$, we get
{\footnotesize 
\begin{align*}
-\eta\left( 1 - \frac{L\eta}{2}\right)   \sum_{t=0}^T\mathbb{E}_{t}\left\|\frac{\partial{F}}{\partial{\mathbf{\delta}^{(1:K)}_{t}}}\right\|^2_F &- \frac{\eta}{2} \sum_{t=0}^T\mathbb{E}_{t}\left\|\frac{\partial{F}}{\partial{\overline{\mathbf{\Gamma}}_t}}\right\|^2_F + \frac{\eta\left( 1 - L\eta\right)}{2} \sum_{t=0}^T\mathbb{E}_{t}\left\|\frac{\partial{F}}{\partial{\mathbf{\Gamma}}_t}\frac{1_n}{n}\right\|^2_F \\
&+ \frac{ \eta L^2}{2n}\sum_{t=0}^T\mathbb{E}_{t}\sum_{i=1}^n\left\|\overline{\mathbf{\Gamma}}_t-\mathbf{\Gamma}_t^{(i)}\right\|^2_F + \frac{L\eta(\sigma_{\Gamma}^2 + \sigma_{\delta}^2)T}{2K}. \numberthis \label{final:eq2} \end{align*}
}

The main challenge however now becomes bounding the term with ${\footnotesize \sum_{t=0}^T\mathbb{E}_{t}\sum_{i=1}^n\left\|\overline{\mathbf{\Gamma}}_t-\mathbf{\Gamma}_t^{(i)}\right\|^2_F }$.  Bounding it requires to derive another lower bound and using an available result. First re-write the $\mathbb{E}\left\|\frac{\partial{f}}{\partial{\mathbf{\Gamma}}_t}\right\|^2_F$ by utilizing Frobenius-norm properties: 
{\footnotesize 
\begin{align*}
\mathbb{E}\left\|\frac{\partial{f}}{\partial{\mathbf{\Gamma}}_t}\right\|^2_F
= & \mathbb{E}\left\|\left(\frac{\partial{f}}{\partial{\mathbf{\Gamma}}_t}-\frac{\partial{F}}{\partial{\mathbf{\Gamma}}_t}\right)+\frac{\partial{F}}{\partial{\mathbf{\Gamma}}_t}\right\|^2_F\\
= & \mathbb{E}\left\|\frac{\partial{f}}{\partial{\mathbf{\Gamma}}_t}-\frac{\partial{F}}{\partial{\mathbf{\Gamma}}_t}\right\|^2_F+\mathbb{E}\left\|\frac{\partial{F}}{\partial{\mathbf{\Gamma}}_t}\right\|^2_F + 2\mathbb{E}\left\langle\frac{\partial{f}}{\partial{\mathbf{\Gamma}}_t}-\frac{\partial{F}}{\partial{\mathbf{\Gamma}}_t},\frac{\partial{F}}{\partial{\mathbf{\Gamma}}_t}\right\rangle\\
= &\mathbb{E}\left\|\frac{\partial{f}}{\partial{\mathbf{\Gamma}}_t}-\frac{\partial{F}}{\partial{\mathbf{\Gamma}}_t}\right\|^2_F+\mathbb{E}\left\|\frac{\partial{F}}{\partial{\mathbf{\Gamma}}_t}\right\|^2_F \end{align*}
}
Then bound each term as follows:
{\footnotesize
\begin{align*}
\leq & n\sigma^2_{\mathbf{\Gamma}} + \mathbb{E}\sum_{i=1}^n\left\|\left(\frac{\partial{F_i}}{\partial{\mathbf{\Gamma}}_t^{(i)}}-\frac{\partial{F_i}}{\partial{\overline{\mathbf{\Gamma}}}_t}\right)+\left(\frac{\partial{F_i}}{\partial{\overline{\mathbf{\Gamma}}}_t} - \frac{\partial{F}}{\partial{\overline{\mathbf{\Gamma}}}_t}\right)+\frac{\partial{F}}{\partial{\overline{\mathbf{\Gamma}}}_t}\right\|^2_F\\
\leq & n\sigma^2_{\Gamma} + \sum_{i=1}^n\mathbb{E}\left\|\frac{\partial{F_i}}{\partial{\mathbf{\Gamma}}_t^{(i)}}-\frac{\partial{F_i}}{\partial{\overline{\mathbf{\Gamma}}}_t}\right\|^2_F + \sum_{i=1}^n\mathbb{E}\left\|\frac{\partial{F_i}}{\partial{\overline{\mathbf{\Gamma}}}_t} - \frac{\partial{F}}{\partial{\overline{\mathbf{\Gamma}}}_t}\right\|^2_F + \mathbb{E}\left\|\frac{\partial{F}}{\partial{\overline{\mathbf{\Gamma}}}_t}\right\|^2_F\\
\leq & n\sigma^2_{\Gamma}+L^2\sum_{i=1}^n\mathbb{E}\left\|\overline{\mathbf{\Gamma}}_t-\mathbf{\Gamma}_t^{(i)}\right\|^2_F+ n\zeta^2+n\mathbb{E}\left\|\frac{\partial{F}}{\partial{\overline{\mathbf{\Gamma}}}_t}\right\|^2_F,
\end{align*}} where first term is bounded by its stochastic gradient variance. To bound the second term $\mathbb{E}\left\|\frac{\partial{F}}{\partial{\mathbf{\Gamma}}_t}\right\|^2_F$ above, first bound the term with sum of the individual components' norm and then use add-subtract trick used with $\frac{\partial{F}}{\partial{\overline{\mathbf{\Gamma}}}_t}$ and $\frac{\partial{F_i}}{\partial{\overline{\mathbf{\Gamma}}}_t}$. Then, use the facts ${\scriptstyle \sum_{i=1}^n\mathbb{E}\left\|\frac{\partial{F_i}}{\partial{\overline{\mathbf{\Gamma}}_t}} - \frac{\partial{F_i}}{\partial{\mathbf{\Gamma}}_t^{(i)}}\right\|^2_F \leq L^2 \mathbb{E}\sum_{i=1}^n\left\|\overline{\mathbf{\Gamma}}_t-\mathbf{\Gamma}_t^{(i)}\right\|^2_F }$(an intermeditate result derived above) and  ${\scriptstyle \sum_{i=1}^n\mathbb{E}\left\|\frac{\partial{F_i}}{\partial{\overline{\mathbf{\Gamma}}}_t} - \frac{\partial{F}}{\partial{\overline{\mathbf{\Gamma}}}_t}\right\|^2_F \leq n\zeta^2}$, while $\zeta$ being the spectral gap of matrix $\mathbf{M}$. Then, from \cite{tang_communication_2018}, we have
{\footnotesize 
\begin{align*}
\sum_{t=1}^{T}\sum_{i=1}^{n}\mathbb{E}\left\|\overline{\mathbf{\Gamma}}_t-\mathbf{\Gamma}_t^{(i)}\right\|^2_F
\leq &   \frac{2}{(1-\zeta)^2}\sum_{t=1}^{T}\eta^2\left\|\frac{\partial{f}}{\partial{\mathbf{\Gamma}}_t}\right\|^2_F,
\end{align*}
}
with the previously derived bound over ${\footnotesize \sum_{t=1}^{T}\eta^2_F\left\|\frac{\partial{f}}{\partial{\mathbf{\Gamma}}_t}\right\|^2_F}$ bound leading to another intermediate bound
{\footnotesize 

\begin{align*}
\left( 1- \frac{2\eta^{2}L^2}{(1-\zeta)^2}  \right)\sum_{t=1}^{T}\sum_{i=1}^{n}\mathbb{E}\left\|\overline{\mathbf{\Gamma}}_t-\mathbf{\Gamma}_t^{(i)}\right\|^2_F
\leq &\frac{2\eta^2}{(1-\zeta)^2}\left(n\sigma^2_{\Gamma} + n\zeta^2\right) +n\sum_{t=1}^{T}\mathbb{E}\left\|\frac{\partial{F}}{\partial{\overline{\mathbf{\Gamma}}}_t}\right\|^2_F\\
\sum_{t=0}^{T}\sum_{i=1}^{n}\mathbb{E}\left\|\overline{\mathbf{\Gamma}}_t-\mathbf{\Gamma}_t^{(i)}\right\|^2_F \leq & C_1\eta^2n\left(\sigma^2_{\Gamma} + \zeta^2 \right)T+\sum_{t=1}^{T}\mathbb{E}\left\|\frac{\partial{F}}{\partial{\overline{\mathbf{\Gamma}}}_t}\right\|^2_F \numberthis\label{final:eq3}
\end{align*}} where $C_1$ being a function of the spectral gap.

Finally, combining \eqref{final:eq2} with the intermediate bound \eqref{final:eq3} together first, bounding the Frobenius norms, and  dividing both sides by $T$ (to average in the end), we get the desired lower bound 
{\footnotesize 
\begin{equation*}
 \frac{2[F(\mathbf{x}_{0})-F_{inf}]}{\eta T} + \frac{\eta L \sigma^{2}}{K} + 
       \eta^{2}L^{2}\sigma^{2} \left( \frac{1+\zeta^{2}}{1-\zeta^{2}}\tau - 1 \right)
\end{equation*}

}
 where $\eta$ is the learning rate that satisfies the given conditions, $ \mathbf{x}_{0} = \{  \overline{\mathbf{\Gamma}}_{0},\mathbf{\delta}^{(1:K)}_{0},\mathbf{\Omega}^{(1:K)}_{0} \}$ is the initial starting point and $ \sigma^2=\sigma^2_{\Gamma}+\sigma^2_{\delta} $ is the total variance bound over stochastic gradients of $\mathbf{\Gamma} $  $\&$ $ \mathbf{\delta}  \square$
 
Our work differs from \cite{wang_cooperative_2018} in that it does not provide adequate theoretical analysis and empirical evaluation for federated learning.

%% file: neurips_2021.bbl
\begin{thebibliography}{75}
\providecommand{\natexlab}[1]{#1}
\providecommand{\url}[1]{\texttt{#1}}
\expandafter\ifx\csname urlstyle\endcsname\relax
  \providecommand{\doi}[1]{doi: #1}\else
  \providecommand{\doi}{doi: \begingroup \urlstyle{rm}\Url}\fi

\bibitem[tox(2017)]{tox21}
Tox21 challenge.
\newblock \url{https://tripod.nih.gov/tox21/challenge/}, 2017.

\bibitem[Ahmed et~al.(2014)Ahmed, Das, and Smola]{ahmed_scalable_2014}
Amr Ahmed, Abhimanyu Das, and Alexander~J. Smola.
\newblock Scalable hierarchical multitask learning algorithms for conversion
  optimization in display advertising.
\newblock In \emph{Proceedings of the 7th {ACM} international conference on
  {Web} search and data mining}, pages 153--162. ACM, 2014.

\bibitem[Baytas et~al.(2016)Baytas, Yan, Jain, and
  Zhou]{baytas_asynchronous_2016}
Inci~M. Baytas, Ming Yan, Anil~K. Jain, and Jiayu Zhou.
\newblock Asynchronous multi-task learning.
\newblock In \emph{Data {Mining} ({ICDM}), 2016 {IEEE} 16th {International}
  {Conference} on}, pages 11--20. IEEE, 2016.

\bibitem[Bottou et~al.(2018)Bottou, Curtis, and
  Nocedal]{bottou_optimization_2018}
L{\'e}on Bottou, Frank~E. Curtis, and Jorge Nocedal.
\newblock Optimization methods for large-scale machine learning.
\newblock \emph{SIAM Review}, 60\penalty0 (2):\penalty0 223--311, 2018.

\bibitem[Caldas et~al.(2018)Caldas, Smith, and
  Talwalkar]{caldas_federated_2018}
Sebastian Caldas, Virginia Smith, and Ameet Talwalkar.
\newblock Federated {Kernelized} {Multi}-{Task} {Learning}.
\newblock \emph{The Conference on Systems and Machine Learning}, page~3, 2018.

\bibitem[Chen et~al.(2020)Chen, Cui, Liu, Wu, and Wang]{chen2020survey}
Chaochao Chen, Jamie Cui, Guanfeng Liu, Jia Wu, and Li~Wang.
\newblock Survey and open problems in privacy preserving knowledge graph:
  Merging, query, representation, completion and applications.
\newblock \emph{arXiv preprint arXiv:2011.10180}, 2020.

\bibitem[Coley et~al.(2017)Coley, Barzilay, Green, Jaakkola, and
  Jensen]{coley2017convolutional}
Connor~W Coley, Regina Barzilay, William~H Green, Tommi~S Jaakkola, and Klavs~F
  Jensen.
\newblock Convolutional embedding of attributed molecular graphs for physical
  property prediction.
\newblock \emph{Journal of chemical information and modeling}, 57\penalty0
  (8):\penalty0 1757--1772, 2017.

\bibitem[Cui et~al.(2014)Cui, Cipar, Ho, Kim, Lee, Kumar, Wei, Dai, Ganger, and
  Gibbons]{cui_exploiting_2014}
Henggang Cui, James Cipar, Qirong Ho, Jin~Kyu Kim, Seunghak Lee, Abhimanu
  Kumar, Jinliang Wei, Wei Dai, Gregory~R. Ganger, and Phillip~B. Gibbons.
\newblock Exploiting {Bounded} {Staleness} to {Speed} {Up} {Big} {Data}
  {Analytics}.
\newblock In \emph{{USENIX} {Annual} {Technical} {Conference}}, pages 37--48,
  2014.

\bibitem[Dai et~al.(2018)Dai, Zhou, Dong, Zhang, and Xing]{dai_toward_2018}
Wei Dai, Yi~Zhou, Nanqing Dong, Hao Zhang, and Eric~P. Xing.
\newblock Toward {Understanding} the {Impact} of {Staleness} in {Distributed}
  {Machine} {Learning}.
\newblock \emph{arXiv:1810.03264 [cs, stat]}, October 2018.
\newblock URL \url{http://arxiv.org/abs/1810.03264}.
\newblock arXiv: 1810.03264.

\bibitem[Dean et~al.()Dean, Corrado, Monga, Chen, Devin, Le, Mao, Ranzato,
  Senior, Tucker, Yang, and Ng]{dean_large_nodate}
Jeffrey Dean, Greg~S Corrado, Rajat Monga, Kai Chen, Matthieu Devin, Quoc~V Le,
  Mark~Z Mao, Marc'Aurelio Ranzato, Andrew Senior, Paul Tucker, Ke~Yang, and
  Andrew~Y Ng.
\newblock Large {Scale} {Distributed} {Deep} {Networks}.
\newblock page~11.

\bibitem[Dutta et~al.(2018)Dutta, Joshi, Ghosh, Dube, and
  Nagpurkar]{dutta_slow_2018}
Sanghamitra Dutta, Gauri Joshi, Soumyadip Ghosh, Parijat Dube, and Priya
  Nagpurkar.
\newblock Slow and {Stale} {Gradients} {Can} {Win} the {Race}:
  {Error}-{Runtime} {Trade}-offs in {Distributed} {SGD}.
\newblock \emph{arXiv preprint arXiv:1803.01113}, 2018.

\bibitem[Duvenaud et~al.(2015)Duvenaud, Maclaurin, Aguilera-Iparraguirre,
  G{\'o}mez-Bombarelli, Hirzel, Aspuru-Guzik, and
  Adams]{duvenaud2015convolutional}
David Duvenaud, Dougal Maclaurin, Jorge Aguilera-Iparraguirre, Rafael
  G{\'o}mez-Bombarelli, Timothy Hirzel, Al{\'a}n Aspuru-Guzik, and Ryan~P
  Adams.
\newblock Convolutional networks on graphs for learning molecular fingerprints.
\newblock \emph{arXiv preprint arXiv:1509.09292}, 2015.

\bibitem[Evgeniou and Pontil(2004)]{evgeniou2004regularized}
Theodoros Evgeniou and Massimiliano Pontil.
\newblock Regularized multi--task learning.
\newblock In \emph{Proceedings of the tenth ACM SIGKDD international conference
  on Knowledge discovery and data mining}, pages 109--117, 2004.

\bibitem[Gilmer et~al.(2017)Gilmer, Schoenholz, Riley, Vinyals, and
  Dahl]{gilmer2017neural}
Justin Gilmer, Samuel~S Schoenholz, Patrick~F Riley, Oriol Vinyals, and
  George~E Dahl.
\newblock Neural message passing for quantum chemistry.
\newblock In \emph{International Conference on Machine Learning}, pages
  1263--1272. PMLR, 2017.

\bibitem[Gupta et~al.(2016)Gupta, Zhang, and Wang]{gupta_model_2016}
Suyog Gupta, Wei Zhang, and Fei Wang.
\newblock Model accuracy and runtime tradeoff in distributed deep learning: {A}
  systematic study.
\newblock In \emph{Data {Mining} ({ICDM}), 2016 {IEEE} 16th {International}
  {Conference} on}, pages 171--180. IEEE, 2016.

\bibitem[Hagberg et~al.(2008)Hagberg, Schult, and Swart]{SciPyProceedings_11}
Aric~A. Hagberg, Daniel~A. Schult, and Pieter~J. Swart.
\newblock Exploring network structure, dynamics, and function using networkx.
\newblock In Ga\"el Varoquaux, Travis Vaught, and Jarrod Millman, editors,
  \emph{Proceedings of the 7th Python in Science Conference}, pages 11 -- 15,
  Pasadena, CA USA, 2008.

\bibitem[Hamilton et~al.(2017)Hamilton, Ying, and Leskovec]{graphsage}
William~L. Hamilton, Rex Ying, and Jure Leskovec.
\newblock Inductive representation learning on large graphs.
\newblock \emph{CoRR}, abs/1706.02216, 2017.
\newblock URL \url{http://arxiv.org/abs/1706.02216}.

\bibitem[He et~al.(2020{\natexlab{a}})He, Annavaram, and
  Avestimehr]{he2020group}
Chaoyang He, Murali Annavaram, and Salman Avestimehr.
\newblock Group knowledge transfer: Federated learning of large cnns at the
  edge.
\newblock \emph{Advances in Neural Information Processing Systems}, 33,
  2020{\natexlab{a}}.

\bibitem[He et~al.(2020{\natexlab{b}})He, Li, So, Zhang, Wang, Wang, Vepakomma,
  Singh, Qiu, Shen, Zhao, Kang, Liu, Raskar, Yang, Annavaram, and
  Avestimehr]{chaoyanghe2020fedml}
Chaoyang He, Songze Li, Jinhyun So, Mi~Zhang, Hongyi Wang, Xiaoyang Wang,
  Praneeth Vepakomma, Abhishek Singh, Hang Qiu, Li~Shen, Peilin Zhao, Yan Kang,
  Yang Liu, Ramesh Raskar, Qiang Yang, Murali Annavaram, and Salman Avestimehr.
\newblock Fedml: A research library and benchmark for federated machine
  learning.
\newblock \emph{arXiv preprint arXiv:2007.13518}, 2020{\natexlab{b}}.

\bibitem[He et~al.(2021)He, Balasubramanian, Ceyani, Rong, Zhao, Huang,
  Annavaram, and Avestimehr]{fedgraphnn}
Chaoyang He, Keshav Balasubramanian, Emir Ceyani, Yu~Rong, Peilin Zhao, Junzhou
  Huang, Murali Annavaram, and Salman Avestimehr.
\newblock Fedgraphnn: {A} federated learning system and benchmark for graph
  neural networks.
\newblock \emph{CoRR}, abs/2104.07145, 2021.
\newblock URL \url{https://arxiv.org/abs/2104.07145}.

\bibitem[Jaggi et~al.(2014)Jaggi, Smith, Tak{\'a}c, Terhorst, Krishnan,
  Hofmann, and Jordan]{jaggi_communication-efficient_2014}
Martin Jaggi, Virginia Smith, Martin Tak{\'a}c, Jonathan Terhorst, Sanjay
  Krishnan, Thomas Hofmann, and Michael~I. Jordan.
\newblock Communication-efficient distributed dual coordinate ascent.
\newblock In \emph{Advances in neural information processing systems}, pages
  3068--3076, 2014.

\bibitem[Jiang et~al.(2020)Jiang, Jung, Karl, and Zhao]{jiang2020federated}
Meng Jiang, Taeho Jung, Ryan Karl, and Tong Zhao.
\newblock Federated dynamic gnn with secure aggregation.
\newblock \emph{arXiv preprint arXiv:2009.07351}, 2020.

\bibitem[Jiang et~al.(2017)Jiang, Balu, Hegde, and
  Sarkar]{jiang_collaborative_2017}
Zhanhong Jiang, Aditya Balu, Chinmay Hegde, and Soumik Sarkar.
\newblock Collaborative deep learning in fixed topology networks.
\newblock In \emph{Advances in {Neural} {Information} {Processing} {Systems}},
  pages 5904--5914, 2017.

\bibitem[Jin et~al.(2016)Jin, Yuan, Iandola, and Keutzer]{jin_how_2016}
Peter~H. Jin, Qiaochu Yuan, Forrest Iandola, and Kurt Keutzer.
\newblock How to scale distributed deep learning?
\newblock \emph{arXiv preprint arXiv:1611.04581}, 2016.

\bibitem[Jin et~al.(2015)Jin, Luo, Zhuang, He, and He]{jin_collaborating_2015}
Xin Jin, Ping Luo, Fuzhen Zhuang, Jia He, and Qing He.
\newblock Collaborating between local and global learning for distributed
  online multiple tasks.
\newblock In \emph{Proceedings of the 24th {ACM} {International} on
  {Conference} on {Information} and {Knowledge} {Management}}, pages 113--122.
  ACM, 2015.

\bibitem[Kairouz et~al.(2019)Kairouz, McMahan, Avent, Bellet, Bennis, Bhagoji,
  Bonawitz, Charles, Cormode, Cummings, et~al.]{kairouz2019advances}
Peter Kairouz, H~Brendan McMahan, Brendan Avent, Aur{\'e}lien Bellet, Mehdi
  Bennis, Arjun~Nitin Bhagoji, Keith Bonawitz, Zachary Charles, Graham Cormode,
  Rachel Cummings, et~al.
\newblock Advances and open problems in federated learning.
\newblock \emph{arXiv preprint arXiv:1912.04977}, 2019.

\bibitem[Kearnes et~al.(2016)Kearnes, McCloskey, Berndl, Pande, and
  Riley]{kearnes2016molecular}
Steven Kearnes, Kevin McCloskey, Marc Berndl, Vijay Pande, and Patrick Riley.
\newblock Molecular graph convolutions: moving beyond fingerprints.
\newblock \emph{Journal of computer-aided molecular design}, 30\penalty0
  (8):\penalty0 595--608, 2016.

\bibitem[Kingma and Ba(2015)]{adam}
Diederik~P. Kingma and Jimmy Ba.
\newblock Adam: A method for stochastic optimization.
\newblock \emph{CoRR}, abs/1412.6980, 2015.

\bibitem[Kipf and Welling(2016)]{kipfgcn}
Thomas~N. Kipf and Max Welling.
\newblock Semi-supervised classification with graph convolutional networks.
\newblock \emph{CoRR}, abs/1609.02907, 2016.
\newblock URL \url{http://arxiv.org/abs/1609.02907}.

\bibitem[Klicpera et~al.(2020)Klicpera, Gro{\ss}, and
  G{\"u}nnemann]{klicpera2020directional}
Johannes Klicpera, Janek Gro{\ss}, and Stephan G{\"u}nnemann.
\newblock Directional message passing for molecular graphs.
\newblock \emph{arXiv preprint arXiv:2003.03123}, 2020.

\bibitem[Kone{\v c}n{\'y} et~al.(2015)Kone{\v c}n{\'y}, McMahan, and
  Ramage]{konecny_federated_2015}
Jakub Kone{\v c}n{\'y}, Brendan McMahan, and Daniel Ramage.
\newblock Federated {Optimization}:{Distributed} {Optimization} {Beyond} the
  {Datacenter}.
\newblock \emph{arXiv:1511.03575 [cs, math]}, November 2015.
\newblock URL \url{http://arxiv.org/abs/1511.03575}.
\newblock arXiv: 1511.03575.

\bibitem[Kone{\v c}n{\'y} et~al.(2016)Kone{\v c}n{\'y}, McMahan, Yu,
  Richt{\'a}rik, Suresh, and Bacon]{konecny_federated_2016}
Jakub Kone{\v c}n{\'y}, H.~Brendan McMahan, Felix~X. Yu, Peter Richt{\'a}rik,
  Ananda~Theertha Suresh, and Dave Bacon.
\newblock Federated {Learning}: {Strategies} for {Improving} {Communication}
  {Efficiency}.
\newblock \emph{arXiv:1610.05492 [cs]}, October 2016.
\newblock URL \url{http://arxiv.org/abs/1610.05492}.
\newblock arXiv: 1610.05492.

\bibitem[Kuhn et~al.(2016)Kuhn, Letunic, Jensen, and Bork]{kuhn2016sider}
Michael Kuhn, Ivica Letunic, Lars~Juhl Jensen, and Peer Bork.
\newblock The sider database of drugs and side effects.
\newblock \emph{Nucleic acids research}, 44\penalty0 (D1):\penalty0
  D1075--D1079, 2016.

\bibitem[Landrum(2006)]{landrum2006rdkit}
Greg Landrum.
\newblock Rdkit: Open-source cheminformatics, 2006.
\newblock URL \url{http://www.rdkit.org}.

\bibitem[Li et~al.(2014)Li, Andersen, Park, Smola, Ahmed, Josifovski, Long,
  Shekita, and Su]{li_scaling_2014}
Mu~Li, David~G. Andersen, Jun~Woo Park, Alexander~J. Smola, Amr Ahmed, Vanja
  Josifovski, James Long, Eugene~J. Shekita, and Bor-Yiing Su.
\newblock Scaling {Distributed} {Machine} {Learning} with the {Parameter}
  {Server}.
\newblock In \emph{{OSDI}}, volume~14, pages 583--598, 2014.

\bibitem[Li et~al.(2018)Li, Han, and Wu]{li2018deeper}
Qimai Li, Zhichao Han, and Xiao-Ming Wu.
\newblock Deeper insights into graph convolutional networks for semi-supervised
  learning.
\newblock In \emph{Proceedings of the AAAI Conference on Artificial
  Intelligence}, volume~32, 2018.

\bibitem[Lian et~al.(2017)Lian, Zhang, Zhang, Hsieh, Zhang, and
  Liu]{lian_can_2017}
Xiangru Lian, Ce~Zhang, Huan Zhang, Cho-Jui Hsieh, Wei Zhang, and Ji~Liu.
\newblock Can {Decentralized} {Algorithms} {Outperform} {Centralized}
  {Algorithms}? {A} {Case} {Study} for {Decentralized} {Parallel} {Stochastic}
  {Gradient} {Descent}.
\newblock \emph{arXiv:1705.09056 [cs, math, stat]}, May 2017.
\newblock URL \url{http://arxiv.org/abs/1705.09056}.
\newblock arXiv: 1705.09056.

\bibitem[Lin et~al.(2018)Lin, Stich, and Jaggi]{lin_dont_2018}
Tao Lin, Sebastian~U. Stich, and Martin Jaggi.
\newblock Don't {Use} {Large} {Mini}-{Batches}, {Use} {Local} {SGD}.
\newblock \emph{arXiv:1808.07217 [cs, stat]}, August 2018.
\newblock URL \url{http://arxiv.org/abs/1808.07217}.
\newblock arXiv: 1808.07217.

\bibitem[Liu et~al.(2017)Liu, Pan, and Ho]{liu_distributed_2017}
Sulin Liu, Sinno~Jialin Pan, and Qirong Ho.
\newblock Distributed multi-task relationship learning.
\newblock In \emph{Proceedings of the 23rd {ACM} {SIGKDD} {International}
  {Conference} on {Knowledge} {Discovery} and {Data} {Mining}}, pages 937--946.
  ACM, 2017.

\bibitem[Ma et~al.(2015)Ma, Smith, Jaggi, Jordan, Richt{\'a}rik, and
  Tak{\'a}{\v c}]{ma_adding_2015}
Chenxin Ma, Virginia Smith, Martin Jaggi, Michael~I. Jordan, Peter
  Richt{\'a}rik, and Martin Tak{\'a}{\v c}.
\newblock Adding vs. {Averaging} in {Distributed} {Primal}-{Dual}
  {Optimization}.
\newblock \emph{arXiv:1502.03508 [cs]}, February 2015.
\newblock URL \url{http://arxiv.org/abs/1502.03508}.
\newblock arXiv: 1502.03508.

\bibitem[Mateos-N{\'u}{\~n}ez et~al.(2015)Mateos-N{\'u}{\~n}ez, Cort{\'e}s, and
  Cortes]{mateos-nunez_distributed_2015}
David Mateos-N{\'u}{\~n}ez, Jorge Cort{\'e}s, and Jorge Cortes.
\newblock Distributed optimization for multi-task learning via nuclear-norm
  approximation.
\newblock In \emph{{IFAC} {Workshop} on {Distributed} {Estimation} and
  {Control} in {Networked} {Systems}}, volume~48, pages 64--69, 2015.

\bibitem[McMahan et~al.(2017)McMahan, Moore, Ramage, Hampson, and
  y~Arcas]{mcmahan2017communication}
Brendan McMahan, Eider Moore, Daniel Ramage, Seth Hampson, and Blaise~Aguera
  y~Arcas.
\newblock Communication-efficient learning of deep networks from decentralized
  data.
\newblock In \emph{Artificial Intelligence and Statistics}, pages 1273--1282.
  PMLR, 2017.

\bibitem[McMahan et~al.(2016{\natexlab{a}})McMahan, Moore, Ramage, Hampson, and
  Arcas]{mcmahan_communication-efficient_2016}
H.~Brendan McMahan, Eider Moore, Daniel Ramage, Seth Hampson, and Blaise
  Ag{\"u}era~y Arcas.
\newblock Communication-{Efficient} {Learning} of {Deep} {Networks} from
  {Decentralized} {Data}.
\newblock \emph{arXiv:1602.05629 [cs]}, February 2016{\natexlab{a}}.
\newblock URL \url{http://arxiv.org/abs/1602.05629}.
\newblock arXiv: 1602.05629.

\bibitem[McMahan et~al.(2016{\natexlab{b}})McMahan, Moore, Ramage, and
  y~Arcas]{fedavg}
H.~Brendan McMahan, Eider Moore, Daniel Ramage, and Blaise~Ag{\"{u}}era
  y~Arcas.
\newblock Federated learning of deep networks using model averaging.
\newblock \emph{CoRR}, abs/1602.05629, 2016{\natexlab{b}}.
\newblock URL \url{http://arxiv.org/abs/1602.05629}.

\bibitem[Mei et~al.(2019)Mei, Guo, Liu, and Pan]{mei2019sgnn}
Guangxu Mei, Ziyu Guo, Shijun Liu, and Li~Pan.
\newblock Sgnn: A graph neural network based federated learning approach by
  hiding structure.
\newblock In \emph{2019 IEEE International Conference on Big Data (Big Data)},
  pages 2560--2568. IEEE, 2019.

\bibitem[Meng et~al.(2021)Meng, Rambhatla, and Liu]{meng2021crossnode}
Chuizheng Meng, Sirisha Rambhatla, and Yan Liu.
\newblock Cross-node federated graph neural network for spatio-temporal data
  modeling, 2021.
\newblock URL \url{https://openreview.net/forum?id=HWX5j6Bv_ih}.

\bibitem[Mitliagkas et~al.(2016)Mitliagkas, Zhang, Hadjis, and
  R{\'e}]{mitliagkas_asynchrony_2016}
Ioannis Mitliagkas, Ce~Zhang, Stefan Hadjis, and Christopher R{\'e}.
\newblock Asynchrony begets momentum, with an application to deep learning.
\newblock In \emph{Communication, {Control}, and {Computing} ({Allerton}), 2016
  54th {Annual} {Allerton} {Conference} on}, pages 997--1004. IEEE, 2016.

\bibitem[Paszke et~al.(2019)Paszke, Gross, Massa, Lerer, Bradbury, Chanan,
  Killeen, Lin, Gimelshein, Antiga, Desmaison, K{\"{o}}pf, Yang, DeVito,
  Raison, Tejani, Chilamkurthy, Steiner, Fang, Bai, and Chintala]{pytorch}
Adam Paszke, Sam Gross, Francisco Massa, Adam Lerer, James Bradbury, Gregory
  Chanan, Trevor Killeen, Zeming Lin, Natalia Gimelshein, Luca Antiga, Alban
  Desmaison, Andreas K{\"{o}}pf, Edward Yang, Zach DeVito, Martin Raison,
  Alykhan Tejani, Sasank Chilamkurthy, Benoit Steiner, Lu~Fang, Junjie Bai, and
  Soumith Chintala.
\newblock Pytorch: An imperative style, high-performance deep learning library.
\newblock \emph{CoRR}, abs/1912.01703, 2019.
\newblock URL \url{http://arxiv.org/abs/1912.01703}.

\bibitem[Ramakrishnan et~al.(2015)Ramakrishnan, Hartmann, Tapavicza, and von
  Lilienfeld]{Ramakrishnan_2015}
Raghunathan Ramakrishnan, Mia Hartmann, Enrico Tapavicza, and O.~Anatole von
  Lilienfeld.
\newblock Electronic spectra from tddft and machine learning in chemical space.
\newblock \emph{The Journal of Chemical Physics}, 143\penalty0 (8):\penalty0
  084111, Aug 2015.
\newblock ISSN 1089-7690.
\newblock \doi{10.1063/1.4928757}.
\newblock URL \url{http://dx.doi.org/10.1063/1.4928757}.

\bibitem[Recht et~al.(2011)Recht, Re, Wright, and Niu]{recht_hogwild:_2011}
Benjamin Recht, Christopher Re, Stephen Wright, and Feng Niu.
\newblock Hogwild: {A} lock-free approach to parallelizing stochastic gradient
  descent.
\newblock In \emph{Advances in neural information processing systems}, pages
  693--701, 2011.

\bibitem[Reddi et~al.(2020)Reddi, Charles, Zaheer, Garrett, Rush,
  Kone{\v{c}}n{\`y}, Kumar, and McMahan]{reddi2020adaptive}
Sashank Reddi, Zachary Charles, Manzil Zaheer, Zachary Garrett, Keith Rush,
  Jakub Kone{\v{c}}n{\`y}, Sanjiv Kumar, and H~Brendan McMahan.
\newblock Adaptive federated optimization.
\newblock \emph{arXiv preprint arXiv:2003.00295}, 2020.

\bibitem[Rogers and Hahn(2010)]{rogers2010extended}
David Rogers and Mathew Hahn.
\newblock Extended-connectivity fingerprints.
\newblock \emph{Journal of chemical information and modeling}, 50\penalty0
  (5):\penalty0 742--754, 2010.

\bibitem[Rohrer and Baumann(2009)]{rohrer2009maximum}
Sebastian~G Rohrer and Knut Baumann.
\newblock Maximum unbiased validation (muv) data sets for virtual screening
  based on pubchem bioactivity data.
\newblock \emph{Journal of chemical information and modeling}, 49\penalty0
  (2):\penalty0 169--184, 2009.

\bibitem[Rong et~al.(2020{\natexlab{a}})Rong, Bian, Xu, Xie, Wei, Huang, and
  Huang]{grover}
Yu~Rong, Yatao Bian, Tingyang Xu, Weiyang Xie, Ying Wei, Wenbing Huang, and
  Junzhou Huang.
\newblock Self-supervised graph transformer on large-scale molecular data,
  2020{\natexlab{a}}.

\bibitem[Rong et~al.(2020{\natexlab{b}})Rong, Bian, Xu, Xie, Wei, Huang, and
  Huang]{rong2020self}
Yu~Rong, Yatao Bian, Tingyang Xu, Weiyang Xie, Ying Wei, Wenbing Huang, and
  Junzhou Huang.
\newblock Self-supervised graph transformer on large-scale molecular data.
\newblock \emph{Advances in Neural Information Processing Systems}, 33,
  2020{\natexlab{b}}.

\bibitem[Rong et~al.(2020{\natexlab{c}})Rong, Xu, Huang, Huang, Cheng, Ma,
  Wang, Derr, Wu, and Ma]{10.1145/3394486.3406474}
Yu~Rong, Tingyang Xu, Junzhou Huang, Wenbing Huang, Hong Cheng, Yao Ma, Yiqi
  Wang, Tyler Derr, Lingfei Wu, and Tengfei Ma.
\newblock Deep graph learning: Foundations, advances and applications.
\newblock In \emph{Proceedings of the 26th ACM SIGKDD International Conference
  on Knowledge DiscoverY; Data Mining}, KDD '20, page 3555–3556, New York,
  NY, USA, 2020{\natexlab{c}}. Association for Computing Machinery.
\newblock ISBN 9781450379984.
\newblock \doi{10.1145/3394486.3406474}.
\newblock URL \url{https://doi.org/10.1145/3394486.3406474}.

\bibitem[Sch{\"u}tt et~al.(2017{\natexlab{a}})Sch{\"u}tt, Arbabzadah, Chmiela,
  M{\"u}ller, and Tkatchenko]{schutt2017quantum}
Kristof~T Sch{\"u}tt, Farhad Arbabzadah, Stefan Chmiela, Klaus~R M{\"u}ller,
  and Alexandre Tkatchenko.
\newblock Quantum-chemical insights from deep tensor neural networks.
\newblock \emph{Nature communications}, 8\penalty0 (1):\penalty0 1--8,
  2017{\natexlab{a}}.

\bibitem[Sch{\"u}tt et~al.(2017{\natexlab{b}})Sch{\"u}tt, Kindermans, Sauceda,
  Chmiela, Tkatchenko, and M{\"u}ller]{schutt2017schnet}
Kristof~T Sch{\"u}tt, Pieter-Jan Kindermans, Huziel~E Sauceda, Stefan Chmiela,
  Alexandre Tkatchenko, and Klaus-Robert M{\"u}ller.
\newblock Schnet: A continuous-filter convolutional neural network for modeling
  quantum interactions.
\newblock \emph{arXiv preprint arXiv:1706.08566}, 2017{\natexlab{b}}.

\bibitem[Shalev-Shwartz and Zhang(2013)]{shalev-shwartz_stochastic_2013}
Shai Shalev-Shwartz and Tong Zhang.
\newblock Stochastic dual coordinate ascent methods for regularized loss
  minimization.
\newblock \emph{Journal of Machine Learning Research}, 14\penalty0
  (Feb):\penalty0 567--599, 2013.

\bibitem[Smith et~al.(2016)Smith, Forte, Ma, Takac, Jordan, and
  Jaggi]{smith_cocoa:_2016}
Virginia Smith, Simone Forte, Chenxin Ma, Martin Takac, Michael~I. Jordan, and
  Martin Jaggi.
\newblock {CoCoA}: {A} {General} {Framework} for {Communication}-{Efficient}
  {Distributed} {Optimization}.
\newblock \emph{arXiv:1611.02189 [cs]}, November 2016.
\newblock URL \url{http://arxiv.org/abs/1611.02189}.
\newblock arXiv: 1611.02189.

\bibitem[Smith et~al.(2017)Smith, Chiang, Sanjabi, and
  Talwalkar]{smith_federated_2017}
Virginia Smith, Chao-Kai Chiang, Maziar Sanjabi, and Ameet~S. Talwalkar.
\newblock Federated multi-task learning.
\newblock In \emph{Advances in {Neural} {Information} {Processing} {Systems}},
  pages 4424--4434, 2017.

\bibitem[Stich(2018)]{stich_local_2018}
Sebastian~U. Stich.
\newblock Local {SGD} {Converges} {Fast} and {Communicates} {Little}.
\newblock September 2018.
\newblock URL \url{https://openreview.net/forum?id=S1g2JnRcFX}.

\bibitem[Suzumura et~al.(2019)Suzumura, Zhou, Baracaldo, Ye, Houck, Kawahara,
  Anwar, Stavarache, Watanabe, Loyola, et~al.]{suzumura2019towards}
Toyotaro Suzumura, Yi~Zhou, Natahalie Baracaldo, Guangnan Ye, Keith Houck, Ryo
  Kawahara, Ali Anwar, Lucia~Larise Stavarache, Yuji Watanabe, Pablo Loyola,
  et~al.
\newblock Towards federated graph learning for collaborative financial crimes
  detection.
\newblock \emph{arXiv preprint arXiv:1909.12946}, 2019.

\bibitem[Tang et~al.(2018)Tang, Gan, Zhang, Zhang, and
  Liu]{tang_communication_2018}
Hanlin Tang, Shaoduo Gan, Ce~Zhang, Tong Zhang, and Ji~Liu.
\newblock Communication {Compression} for {Decentralized} {Training}.
\newblock \emph{arXiv:1803.06443 [cs, stat]}, March 2018.
\newblock URL \url{http://arxiv.org/abs/1803.06443}.
\newblock arXiv: 1803.06443.

\bibitem[Veličković et~al.(2018)Veličković, Cucurull, Casanova, Romero,
  Liò, and Bengio]{gat}
Petar Veličković, Guillem Cucurull, Arantxa Casanova, Adriana Romero, Pietro
  Liò, and Yoshua Bengio.
\newblock Graph attention networks, 2018.

\bibitem[Wang et~al.(2020)Wang, Li, Li, and Chen]{wang2020graphfl}
Binghui Wang, Ang Li, Hai Li, and Yiran Chen.
\newblock Graphfl: A federated learning framework for semi-supervised node
  classification on graphs.
\newblock \emph{arXiv preprint arXiv:2012.04187}, 2020.

\bibitem[Wang et~al.(2016)Wang, Kolar, and Srerbo]{wang_distributed_2016}
Jialei Wang, Mladen Kolar, and Nathan Srerbo.
\newblock Distributed multi-task learning.
\newblock In \emph{Artificial {Intelligence} and {Statistics}}, pages 751--760,
  2016.

\bibitem[Wang and Joshi(2018)]{wang_cooperative_2018}
Jianyu Wang and Gauri Joshi.
\newblock Cooperative {SGD}: {A} unified {Framework} for the {Design} and
  {Analysis} of {Communication}-{Efficient} {SGD} {Algorithms}.
\newblock August 2018.
\newblock URL \url{https://arxiv.org/abs/1808.07576v2}.

\bibitem[Wu et~al.(2018)Wu, Ramsundar, Feinberg, Gomes, Geniesse, Pappu,
  Leswing, and Pande]{wu2018moleculenet}
Zhenqin Wu, Bharath Ramsundar, Evan~N Feinberg, Joseph Gomes, Caleb Geniesse,
  Aneesh~S Pappu, Karl Leswing, and Vijay Pande.
\newblock Moleculenet: a benchmark for molecular machine learning.
\newblock \emph{Chemical science}, 9\penalty0 (2):\penalty0 513--530, 2018.

\bibitem[Yang et~al.(2019)Yang, Swanson, Jin, Coley, Eiden, Gao, Guzman-Perez,
  Hopper, Kelley, Mathea, et~al.]{yang2019analyzing}
Kevin Yang, Kyle Swanson, Wengong Jin, Connor Coley, Philipp Eiden, Hua Gao,
  Angel Guzman-Perez, Timothy Hopper, Brian Kelley, Miriam Mathea, et~al.
\newblock Analyzing learned molecular representations for property prediction.
\newblock \emph{Journal of chemical information and modeling}, 59\penalty0
  (8):\penalty0 3370--3388, 2019.

\bibitem[Yang(2013)]{yang_trading_2013}
Tianbao Yang.
\newblock Trading computation for communication: {Distributed} stochastic dual
  coordinate ascent.
\newblock In \emph{Advances in {Neural} {Information} {Processing} {Systems}},
  pages 629--637, 2013.

\bibitem[Yang et~al.(2013)Yang, Zhu, Jin, and Lin]{yang_analysis_2013}
Tianbao Yang, Shenghuo Zhu, Rong Jin, and Yuanqing Lin.
\newblock Analysis of distributed stochastic dual coordinate ascent.
\newblock \emph{arXiv preprint arXiv:1312.1031}, 2013.

\bibitem[Yu et~al.(2018)Yu, Yang, and Zhu]{yu_parallel_2018}
Hao Yu, Sen Yang, and Shenghuo Zhu.
\newblock Parallel {Restarted} {SGD} with {Faster} {Convergence} and {Less}
  {Communication}: {Demystifying} {Why} {Model} {Averaging} {Works} for {Deep}
  {Learning}.
\newblock \emph{arXiv:1807.06629 [cs, math]}, July 2018.
\newblock URL \url{http://arxiv.org/abs/1807.06629}.
\newblock arXiv: 1807.06629.

\bibitem[Zhang and Yeung(2012)]{zhang_convex_2012}
Yu~Zhang and Dit-Yan Yeung.
\newblock A convex formulation for learning task relationships in multi-task
  learning.
\newblock \emph{arXiv preprint arXiv:1203.3536}, 2012.

\bibitem[Zhou et~al.(2020)Zhou, Chen, Zheng, Zheng, Wu, Liu, and
  Wang]{zhou2020privacy}
Jun Zhou, Chaochao Chen, Longfei Zheng, Xiaolin Zheng, Bingzhe Wu, Ziqi Liu,
  and Li~Wang.
\newblock Privacy-preserving graph neural network for node classification.
\newblock \emph{arXiv preprint arXiv:2005.11903}, 2020.

\end{thebibliography}
